\documentclass[10pt,twocolumn,letterpaper]{article}

\usepackage[pagenumbers]{cvpr}

\usepackage{amsfonts}
\usepackage{amsmath}
\usepackage{amsthm}
\usepackage{bm}
\usepackage{multirow}
\usepackage{graphicx}
\usepackage{enumitem}
\usepackage{array}
\usepackage{wrapfig}
\usepackage{xcolor}
\usepackage{shadethm}
\usepackage{thmtools}
\usepackage{algorithm}
\usepackage{algpseudocode}
\usepackage{nicefrac}
\usepackage{microtype}
\usepackage{booktabs}
\usepackage[para,online,flushleft]{threeparttable}
\usepackage{pifont}
\usepackage[labelfont=bf,textfont=it]{caption}

\def\1{\bm{1}}

\DeclareMathAlphabet{\mathsfit}{\encodingdefault}{\sfdefault}{m}{sl}
\SetMathAlphabet{\mathsfit}{bold}{\encodingdefault}{\sfdefault}{bx}{n}

\newtheorem{lemma}{Lemma}

\newtheorem{definition}{Definition}
\newtheorem{remark}{Remark}
\newtheorem{assumption}{Assumption}

\declaretheoremstyle[%
  spaceabove=-6pt,%
  spacebelow=6pt,%
  headfont=\bfseries\itshape,%
  postheadspace=0.5em,%
  qed=\qedsymbol%
]{mystyle} 

\declaretheorem[
  shaded={rulecolor=black, rulewidth=1pt, bgcolor=blue!5!white},
  name=Theorem,
]{theo}

\theoremstyle{mystyle}

\definecolor{cvprblue}{rgb}{0.21,0.49,0.74}
\usepackage[pagebackref,breaklinks,colorlinks,allcolors=cvprblue]{hyperref}
\newcommand{\colref}[3]{\hyperref[#2]{#1~\ref*{#2}{#3}}}
\newcommand{\figref}[1]{\colref{Figure}{#1}{}}

\newcommand{\eqnref}[1]{\colref{Eq.}{#1}{}}
\newcommand{\tabref}[1]{\colref{Table}{#1}{}}
\newcommand{\theoremref}[1]{\colref{Theorem}{#1}{}}
\newcommand{\assmref}[1]{\colref{Assumption}{#1}{}}
\newcommand{\lemmaref}[1]{\colref{Lemma}{#1}{}}

\newcommand{\cmark}{\textcolor{green!80!black}{\ding{51}}}
\newcommand{\xmark}{\textcolor{red}{\ding{55}}}

\def\paperID{*****}
\def\confName{CVPR}
\def\confYear{2025}

\title{STITCH: Surface reconstrucTion using Implicit neural representations with Topology Constraints and persistent Homology}

\author{
  \textbf{Anushrut Jignasu}$^1$ \quad
  \textbf{Ethan Herron}$^1$ \quad
  \textbf{Zhanhong Jiang}$^1$ \quad
  \textbf{Soumik Sarkar}$^1$ \quad 
  \textbf{Chinmay Hegde}$^2$ \\
  \textbf{Baskar Ganapathysubramanian}$^1$ \quad
  \textbf{Aditya Balu}$^{1}$ \quad
  \textbf{Adarsh Krishnamurthy}$^{1*}$ \\[1ex]
  $^1$Iowa State University \quad $^2$New York University \quad $^*$Corresponding author\\[1ex]
}

\begin{document}
\maketitle

\begin{abstract}
We present STITCH, a novel approach for neural implicit surface reconstruction of a sparse and irregularly spaced point cloud while enforcing topological constraints (such as having a single connected component). We develop a new differentiable framework based on persistent homology to formulate topological loss terms that enforce the prior of a single 2-manifold object. Our method demonstrates excellent performance in preserving the topology of complex 3D geometries, evident through both visual and empirical comparisons. We supplement this with a theoretical analysis, and provably show that optimizing the loss with stochastic (sub)gradient descent leads to convergence and enables reconstructing shapes with a single connected component. Our approach showcases the integration of differentiable topological data analysis tools for implicit surface reconstruction.
\end{abstract}

\section{Introduction}
\label{intro}

Surface reconstruction from point clouds is an extensively studied problem in computer vision and graphics. The goal is to reconstruct the underlying surface from a sampled set of 3D points of a point cloud. Conventional surface reconstruction approaches can be categorized as \emph{implicit} or \emph{explicit}: explicit methods use explicit triangulation or parametric splines to define the surface, while implicit methods learn the \emph{level set}, $W = \{x \in \mathbb{R}^3 | f(x) = z\}$ of some parameterized function $f: \mathbb{R}^3 \rightarrow \mathbb{R}$. Typically, the zero-level set ($z = 0$) is utilized for extracting the surface. 

\begin{figure}[t!]
    \centering
    \includegraphics[width=0.88\linewidth,trim={1.5in 2.5in 1.5in 2.5in},clip]{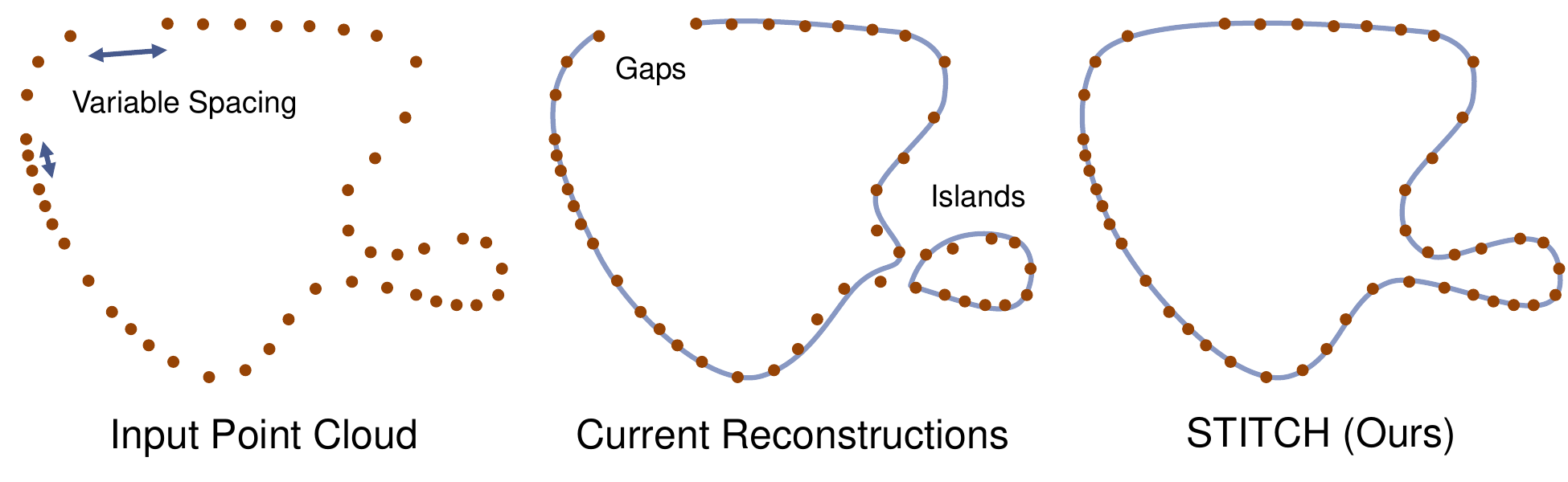}
    \caption{Current reconstruction approaches prioritize accuracy of the reconstruction over the topology of the reconstructed object. This leads to cases where the method has to be tuned for the point cloud spacing, otherwise leading to gaps or isolated islands in the reconstruction. In our approach, we apply topological constraints to obtain a single connected 2-manifold surface.}
    \label{fig:overview}
\end{figure}

Current state-of-the-art approaches have used neural networks with implicit representations to develop a class of methods called implicit neural representations (INR)~\citep{sitzmann2020implicit,atzmon2020sal,gropp2020implicit}. These approaches leverage a multi-layer perceptron (MLP) architecture and utilize the MLP's inductive bias to represent smooth surfaces. They mathematically formulate the surface reconstruction task by using a continuous parameterized function to map the domain of the input (usually a point cloud) to a distance space~\citep{chen2019learning,chibane2020implicit,tagliasacchi2009curve}. However, the focus of current reconstruction methods has been to reduce the error between the reconstructed surface and the input point cloud or to obtain a smooth reconstruction. These reconstructions are not guaranteed to maintain connected components, especially for inputs consisting of thin or sparsely sampled regions (see \figref{fig:overview}). In addition, ensuring a geometry without spurious disconnected components or islands remains a significant research challenge. Practical applications---such as performing physical simulations or volumetric meshing of physical objects---require a watertight 2-manifold surface mesh to be reconstructed from the point cloud, which can only be enforced with topological constraints.

Topological data analysis (TDA) provides topological descriptors for a given input data (such as a point cloud). The computations underlying TDA are often built using tools from \textit{persistent homology}. The topological features are iteratively defined using a process called \textit{filtration} and visualized using \textit{persistence diagrams}. For surface reconstruction, \emph{0-dimensional features} of \textit{persistence diagrams} are of particular interest as they represent features of the input data that contribute to a single connected component. In this work, we leverage this specific topological descriptor to guide the loss function of our neural implicit surface reconstruction to generate a single connected component. 

\begin{figure}[t!]
    \centering
    \includegraphics[width=0.99\linewidth,trim={0.0in 1.6in 0.0in 1.6in},clip]{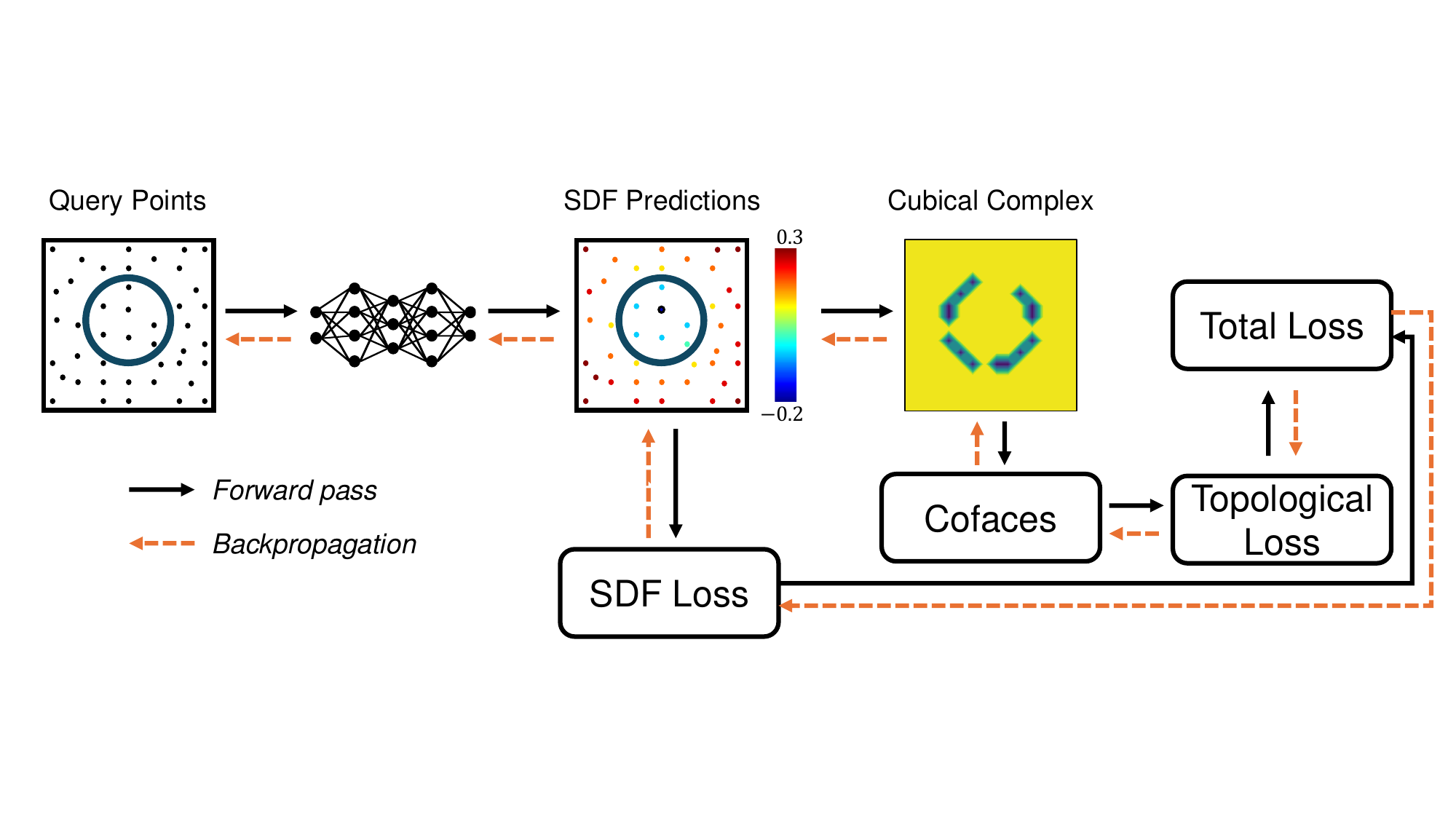}
    \caption{We propose a neural implicit surface reconstruction approach, STITCH, that leverages persistent homology and builds upon Neural-Pull~\citep{ma2020neural}. We begin with a point cloud, predict its signed distance field through an implicit neural representation, and compute a cubical complex. Through the filtration process, we generate various topological features, which are penalized using our differentiable topological loss terms. The topological losses encourage significant features to persist and simultaneously diminish the persistence of noisy features, leading to a single connected component in the reconstruction.}
    \label{fig:STITCHOverview}
\end{figure}

We show that incorporating this descriptor as an additional loss term in the INR surface reconstruction enables the generation of a single-connected surface from sparse and irregularly spaced point cloud data (\figref{fig:STITCHOverview}). Recently, \citet{papamarkou2024position} published a position paper postulating that topological deep learning is the new frontier for relational learning and can complement geometric deep learning by incorporating topological concepts. To the best of our knowledge, we are the first to integrate such topological constraints for neural implicit surface reconstruction. To summarize, the main contributions of our work include the following:
\begin{enumerate}
    \item A novel differentiable topological loss term that enforces topological constraints for neural implicit surface reconstruction. This new loss can be seamlessly used in conjunction with standard INR surface reconstruction approaches. We refer the reader to \cref{tab:comparison} for an overview of methods that contribute to aspects of proposed approach.
    \item Theoretical analysis showing that optimizing this combined loss with stochastic (sub)gradient descent guarantees convergence to a solution that possesses a single connected component.
    \item Practical demonstration of our method with several numerical experiments to show that integrating such TDA tools enables the reconstruction of a single connected 2-manifold mesh, especially when dealing with relatively sparse point clouds. Note that our proposed approach is aimed towards topological feature preservation as opposed to reconstruction accuracy.
\end{enumerate}

\section{Related work}
\label{related_work}
\textbf{Surface Reconstruction:} The problem of reconstructing surfaces from point clouds has been investigated for many years. Please see the Supplement for a more thorough description of surface reconstruction methods. The main advantage of implicit methods is that they handle noise in the data much better than pure geometric approaches. Leveraging the generalized function approximation nature of neural networks has recently allowed the replacement of the implicit function with a neural network under the class of \emph{Implicit Neural Representations}~\citep{atzmon2020sal,atzmon2021sald,liu2022learning,ben2021digs}. Specifically, if the implicit function is the distance field, it allows for the use of physics-informed methods that solve the Eikonal equation using a boundary value problem for surface reconstruction~\citep{sitzmann2020implicit,gropp2020implicit}. This approach guides the distance function to have a value of zero on the underlying surface represented by the point cloud. \tabref{tab:comparison} provides a concise comparison of some of the current state-of-the-art surface reconstruction approaches. In this paper, we propose a neural implicit surface reconstruction approach that incorporates topological constraints during the training process. 

\begin{table}[!t]
\centering
\small
\caption{An overview of various works leveraging persistent homology and/or implicit surface reconstruction and their corresponding contributions. Key - Topology-controllable Implicit Surface Reconstruction Based on Persistent Homology (TCISR), Multiparameter Persistent Homology (MPH), Connectivity-Optimized Representation Learning via Persistent Homology (COPH), Differentiability and Optimization of Multiparameter Persistent Homology (DOMPH), Geometry-Informed Neural Networks (GINN) and Implicit Neural Representation (INR).}
\setlength{\tabcolsep}{3pt}  
\setlength\extrarowheight{2pt}
\begin{tabular}{@{}l @{}c c c c@{}}  
    \toprule
    \textbf{Method} & \textbf{Unified loss} & \textbf{Convergence} & \textbf{Connectivity} & \textbf{INR}\\ \midrule
      \texttt{NP~\citep{ma2020neural}}  & \xmark& \xmark  & \xmark &\cmark\\
      \texttt{TCISR~\citep{dong2022topology}}   & \cmark  &  \xmark  &\cmark  &   \xmark    \\ 
      \texttt{MPH~\citep{loiseaux2024stable}}  & \xmark & \cmark  &  \cmark    &   \xmark    \\
      \texttt{COPH~\citep{hofer2019connectivity}}   & \cmark& \xmark  & \cmark   &   \xmark   \\
      \texttt{DOMPH~\citep{scoccola2024differentiability}} & \xmark & \cmark & \cmark & \xmark \\ 
      \texttt{GINN~\citep{berzins2024geometry}} & \cmark & \xmark & \cmark & \cmark \\ \midrule
      \texttt{\textbf{STITCH}} (Ours)   & \cmark& \cmark   &  \cmark     &   \cmark  \\
      \bottomrule
\end{tabular}
\vspace{-0.1in}
\label{tab:comparison}
\end{table}

\noindent\textbf{Topological Surface Reconstruction:} Topological machine learning~\citep{hensel2021survey, bruel2019topology} has recently emerged as a powerful fusion of two fields---\textit{topological data analysis} and \textit{machine learning}. Topological data analysis (TDA) provides insights relating to algebraic topology---invariant properties of spaces under continuous transformations. We refer the reader to \citet{hensel2021survey} for a more in-depth survey of TDA. Persistent Homology (PH) is the primary tool used in TDA for generating multi-scale insights about topological criteria such as \textit{connected components}, \textit{loops}, or \textit{voids}. These features are often referred to as \textit{0-dimensional}, \textit{1-dimensional}, or \textit{2-dimensional} features, respectively. Controlling topology during the surface reconstruction process has gained attention in recent years. 

Approaches leveraging PH have been used for shape description and classification~\citep{carlsson2005persistence}, mesh segmentation~\citep{skraba2010persistence}, and more related to our work, surface reconstruction~\citep{chazal2008towards, bruel2020topology}. \citet{bruel2020topology} incorporated topological priors to obtain a likelihood function over the reconstruction domain. PH has also been leveraged for 3D reconstruction from 2D images by introducing a loss term that penalizes the topological features of a predicted 3D shape using 3D ground truth-based topological features. \citet{mezghanni2021physically} also propose a similar approach in using generative networks for shape generation. \citet{dong2022topology} proposed a target function using persistent pairs in the persistent diagram to control the coefficients of a B-spline function. 

Conventional neural implicit surface reconstruction approaches often suffer from the well-known problem of generating ghost geometries away from the underlying surface. This is primarily due to off-surface points being close enough to each other, causing the neural network to generate a signed distance field value of zero for such points. As a result, this generates more than one single connected component. Although loss terms penalizing such a scenario have been proposed in other works~\citep{sitzmann2020implicit, pmlr-v139-lipman21a}, they do not explicitly enforce a penalty on multiple components. In this work, we propose leveraging persistent homology for gathering and utilizing topological information relating to the number of connected components. We build our approach on a neural implicit representation framework similar to Neural-Pull~\citep{ma2020neural}, with added topological losses that enforce a single connected component~\citep{dong2022topology} in the reconstructed mesh. We start with a relatively sparse point cloud and predict a signed distance function without requiring ground truth normal information.

Topological information for surface reconstruction from a point cloud can be incorporated by three different approaches. (i) Given a point cloud, compute the alpha complex or rips complex and use the resultant birth and death times to filter out unwanted features. This is followed by a conventional reconstruction approach such as Poisson surface reconstruction. (ii) Given a point cloud, reconstruct a mesh using a conventional method. Compute the topological features by sampling points on the mesh and use the alpha/rips complex to refine the point set and reconstruct a final mesh. (iii) Operate in the implicit space and compute a distance function for a point cloud. Often, this distance function is computed over a grid of points, making it suitable for a cubical complex-based topological feature computation as opposed to alpha/rips complex computed from point clouds. Use the computed features to penalize the SDF and reconstruct a mesh using marching cubes. We build upon the last approach in this work for the following reasons: (i) SDF provides a continuous and smooth approximation of the underlying shape, capturing fine geometric details that discrete point samples might miss; (ii) the use of topological constraints in the implicit space has been under-explored, and (iii) this approach allows us to build an end-to-end differentiable neural implicit surface reconstruction pipeline that incorporates topological features.

\section{Our Method: STITCH}
\label{method}

Our approach leverages topological data analysis, particularly persistent homology, to capture the evolution of the total number of connected components as a function of a scale parameter. We build on the formulation of Neural-Pull~\citep{ma2020neural} to learn the signed distance function (SDF). However, we note that our approach can be added to any existing neural implicit surface reconstruction method.

\subsection{Problem formulation and preliminaries}
\label{prelim}
We formulate the surface reconstruction as an optimization problem minimizing a unified loss that consists of {reconstruction} and {connectivity} losses, respectively. Intuitively, the reconstruction loss establishes the topological surface from the given 3D point cloud, while the connectivity loss encourages connected components throughout the surface.

We first denote by $\mathbf{P}=\{\mathbf{P}_i\}_{i=1}^L$ a 3D point cloud, where $\mathbf{P}_i=(x_i,y_i,z_i)$, $N$ is the total number of points, given $\mathbf{P}\subset\mathbb{R}^3$. SDF is denoted by a nonlinear mapping $f:\mathbb{R}^3\to\mathbb{R}$ such that it can be parameterized by some parameters $\theta\in\mathbb{R}^d$. SDF is essentially {learned by pulling a 3D query location $\mathbf{q}$, which is randomly sampled around a 3D point $\mathbf{P}_i$, to its nearest neighbor $\mathbf{c}$ on the topological surface}. Denote by $\mathbf{Q}=\{\mathbf{q}_j\}_{j=1}^W$ the set of query points such that pulling these points requires a direction and a step size. We utilize the gradient of $f$ respective to $\theta$ at the query point $\mathbf{q}_j$ (either positive or negative), intuitively rendering us the fastest direction of change along the SDF landscape in 3D space. The gradient can be computed correspondingly through backpropagation during training. Mathematically, it can be expressed by:
{\small
\begin{equation}
\label{np_loss}
    \mathbf{c}'_j=\mathbf{q}_j-f_{\mathbf{q}_j} \mathbf{g}_j,
\end{equation}
}where $\mathbf{c}'_j$ is the predicted pulled location (nearest neighbor) of $\mathbf{q}_j$, $f_{\mathbf{q}_j}$ indicates a SDF value calculated by $f$ at the query point $\mathbf{q}_j$, and $\mathbf{g}_j=\frac{\nabla_\theta f_{\mathbf{q}_j}}{\|\nabla_\theta f_{\mathbf{q}_j}\|}$ is the direction, where $\|\cdot\|$ is the Euclidean norm. In this context, $f_{\mathbf{q}_j}$ acts like a step size in \eqnref{np_loss}, which resembles a normalized SGD update.
The goal of Neural-Pull is to learn to ``pull'' each $\mathbf{q}_j\in\mathbf{Q}$ to its corresponding $\mathbf{c}_j$ on the point cloud. To this end, we minimize the distance between the predicted location $\mathbf{c}'_j$ and the nearest neighbor $\mathbf{c}_j$ on the topological surface, which can be represented by:
{\small
\begin{equation}
\label{loss_sdf}
    \mathcal{L}_{g}(\theta;\mathbf{P})=\frac{1}{W}\sum_{j=1}^W\|\mathbf{c}'_j-\mathbf{c}_j\|^2.
\end{equation}
}In $\mathcal{L}_{g}(\theta;\mathbf{P})$, we hide $\mathbf{Q}$ since its points are randomly sampled around the points from $\mathbf{P}$. \eqnref{loss_sdf} signifies the MSE between the predicted pulled locations of all query points and their corresponding nearest neighbors. The surface reconstruction is attained by minimizing the above equation; however, there is no expectation that the reconstructed surface is connected, which is mitigated by defining topological losses via persistent homology theory.

\subsection{Basics of persistent homology}
Persistent homology is a method broadly used to infer multi-scale topological features from a 3D point cloud (or a discrete space filtered by a real-valued function). Common features include connected components, loops, and voids; in this work, our focus is primarily on connected components while noting that the technique can plausibly be extended to other features. We briefly review key concepts. 

\noindent\textbf{Cubical complex.} A cubical complex~\citep{wang2016object,bleile2022persistent} is a discretization of space built from different elements, such as vertices (0-cubes), line segments (1-cubes), squares (2-cubes), and spatial cubes (3-cubes). For a cubical complex, any faces of a cube are included in it, which means for any 1-cubes, the two endpoints of the line segment are contained in the complex. Thus, the union of all cubes forms correspondingly the underlying space of the complex such that a cubical complex can be regarded as a technique to discretize its underlying space. In terms of implementation, cubical complexes are relatively easy to establish and suitable for processing grid-like data like images and voxels~\citep{dong2022topology,hu2024topology,waibel2022capturing}.

\noindent\textbf{Filtration.} To mathematically compute and represent the topological features, homology is defined over a cubical complex $\mathcal{C}$ to calculate $k^{th}$ homology group $\mathcal{H}_k(\mathcal{C})$ such that $k$-dimensional topological features in $\mathcal{C}$ can be effectively encoded, including connected components, loops, and voids. Additionally, persistent homology manipulates a sequence of nested complexes and captures the evolution of topological features in the sequence. The construction and destruction of topological features are known as \textit{birth} and \textit{death}, respectively. These two values are typically adopted to signify a topological feature, denoted as $(b,d)$. Hence, the union of all these paired values is called the persistence diagram (PD), and the corresponding absolute value $|b-d|$ is the persistence of the topological feature. We refer to the appendix for an example of the filtration process. In this work, we use the SDF as a filtration function to determine the birth and death times in the PD. However, establishing the connection, particularly in the backpropagation between the topological features and SDF, is non-trivial, particularly for maintaining the end-to-end differentiability as shown in \figref{fig:overview}. We will resort to a key concept of \textit{coface} to mitigate the issue. It should be noted that though $d>b$ such that $|b-d|$ can be rewritten as $d-b$, we use $|b-d|$ as our notation to keep in line with the style used by previous related work~\citep{dong2022topology}.

\subsection{Incorporating topological constraints}
\label{top_loss}
Our goal in using TDA in this work is to obtain a single connected component of the object in the SDF space. However, we note that our approach is general and can incorporate more complex topological outcomes. Our topological loss function is based on the following assumptions (inspired from \citet{dong2022topology} and \citet{hu2024topology}).

\begin{figure}
    \centering
    \includegraphics[width=0.9\linewidth,trim={1.2in 2.0in 0.0in 1.25in},clip]{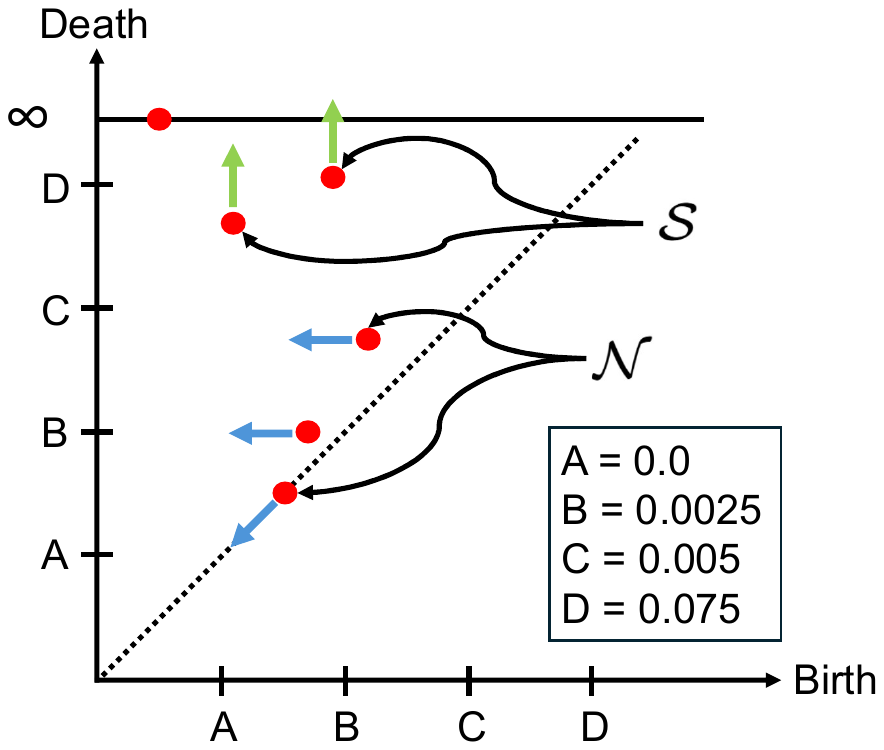}
    \caption{A visual representation of topological features represented by the persistence diagram. The arrows on the persistence diagram showcase the direction of movement for each set of features. Features with green arrows are considered significant ($\mathcal{S}$), and we want to preserve them, while features with a light blue arrow are considered as noise ($\mathcal{N}$), and we minimize them.}
    \label{fig:loss_formulation}
\end{figure}

The points on the PD that are close to the diagonal having equal birth and death times are considered \textit{noise} features. We want to minimize their persistence. For topological features to appear on the zero-level set, the birth times of \textit{significant} features (i.e. features having large persistence) should be minimized. The points on PD having large persistence should persist across scales, otherwise reduce their scale. We illustrate the effect on these features in \figref{fig:loss_formulation}. To ease the analysis, denote by $\mathcal{N}=\{(b_i,d_i)|b_i=d_i,i\in\mathcal{I}_n\}$ and $\mathcal{S}=\{(b_i,d_i)|d_i>b_i,i\in\mathcal{I}_s\}$ the noisy feature set and the significant feature set, where $\mathcal{I}_n$ is the index set that numbers some features close to the diagonal, $\mathcal{I}_s$ is the index set that numbers some features are significant. Set $|\mathcal{N}|=N$ and $|\mathcal{S}|=S$ and denote by $\mathcal{L}_{\mathcal{N}}$ and $\mathcal{L}_{\mathcal{S}}$ the induced noisy loss and significant loss,
such that the following equations are obtained:
{
\small
\begin{equation}\label{connectivity_loss}
    \mathcal{L}_{\mathcal{N}} = \sum_{i=1}^{N} b_i + \sum_{j=1}^{N} (d_j - b_j),\;
    \mathcal{L}_{\mathcal{S}} = -\sum_{i=1}^{S} (d_i - b_i)
\end{equation}
}
As $f_\theta$ is used as a filtration function to obtained $b_i$ and $d_i$,
by denoting the connectivity loss as $\mathcal{L}_c(\theta;\mathbf{P}):\mathbb{R}^3\times\mathbb{R}^d\to\mathbb{R}$, it is immediately obtained that $\mathcal{L}_c(\theta;\mathbf{P})=\lambda_1\mathcal{L}_{\mathcal{S}} + \lambda_2\mathcal{L}_{\mathcal{N}}$. Detailed values for $\lambda_1$ and $\lambda_2$ are specified in Appendix.

\noindent Hence, the optimization problem for the proposed STITCH can be framed by minimizing the following unified loss: 
{\small
\begin{equation}\label{eq_loss}
    \textnormal{min}_{\theta}\mathcal{L}(\theta;\mathbf{P})=\mathcal{L}_{g}(\theta;\mathbf{P})+\mathcal{L}_c(\theta;\mathbf{P}),
\end{equation}}
Optimizing $\theta$ in the above equation will have an impact on the topological features. Denote by $\theta^*$ the optimal parameters such that the optimal topological features $(b^*_i, d^*_i)$ can be determined immediately. We remark the significance of $\mathcal{L}(\theta;\mathbf{P})$ on two folds. On the one hand, this is the first attempt to combine the favored neural implicit representation and TDA in a unified loss through an established end-to-end framework, which tactically ensures the optimal 3D surface reconstruction with connected components. On the other hand, though connectivity loss has been introduced in~\citep{dong2022topology}, their surface reconstruction is through using B-spline function approximation, inevitably resulting in several potential issues, including how to determine the number of knots and the interpolating oscillation. However, the neural implicit representation is entirely data-driven to search for the optimal parameters for SDF. Through the tuning of coefficients $\lambda_1$ and $\lambda_2$, the additional connectivity loss imposes a constraint for the surface reconstruction to guarantee the connectivity, which we validated through our empirical outcomes. So far, we have known that $b_i$ and $d_i$ are calculated by using $f_\theta$ in the forward pass, but in the backpropagation, how to ensure the differentiability remains unclear. Modern deep-learning packages rely on automatic differentiation when computing gradients. Thus, in what follows, we reveal its end-to-end differentiability to set the foundation for the implementation.

\subsection{Differentiability}
\label{differentiable_tda}
To ensure the differentiability in the pipeline shown in \figref{fig:overview}, we first introduce a key definition for coface.
\begin{definition}
    A coface in a cubical complex represents a higher-dimensional cell that contains a given cell as part of its boundary. For any $\varsigma\in\mathcal{C}$, its non-empty subsets $\tau\subset\varsigma$ are called faces of $\varsigma$, and $\varsigma$ is called a coface of $\tau$.
\end{definition}
For instance, $[0,1]$ is a face of $[0,1,2]$, and $[0,1,2]$ is a coface of $[0,1]$. Particularly, for a birth event in this study, the coface represents the cell whose addition created a topological feature. Analogously, for a death event, the coface represents the cell whose addition terminated a topological feature. 
In light of the above definition, the coface of a topological feature can be used to assist in the gradient computation of $\mathcal{L}_c$ with respect to $\theta$ in the backpropagation as it stores the indices of birth and death events associated with that feature, from which the SDF values are fetched accordingly. To help with the illustration of how to achieve this, we denote by $\mathcal{H}_b\circ f_\theta$ and $\mathcal{H}_{b,d}\circ f_\theta$ the \textit{birth and persistence functionals}. Without loss of generality, we drop $j$ in $\mathbf{c}'_j$ and $\mathbf{g}_j$ and equip $\mathbf{g}$ with $\theta$ for the analysis.

We calculate the gradients of individual components of $\mathcal{L}(\theta;\mathbf{P})$ in the following.
{
\small
\begin{equation}\label{geo_grad}
    \frac{\partial \mathcal{L}_g}{\partial\theta}=\frac{\partial \mathcal{L}_g}{\partial \mathbf{c}'}\bigg[\frac{\partial \mathbf{c}'}{\partial f_\theta}\frac{\partial f_\theta}{\partial \theta}+\frac{\mathbf{c}'}{\partial \mathbf{g}_\theta}\frac{\partial \mathbf{g}_\theta}{\partial\theta}\bigg]
\end{equation}
\begin{equation}\label{sig_grad}
    \frac{\partial \mathcal{L}_\mathcal{S}}{\partial\theta}=-\sum_{i}^S\bigg[\frac{\partial (\mathcal{H}_{b,d}\circ f^i_\theta)}{\partial f^i_\theta}\frac{\partial f^i_\theta}{\partial \theta}\bigg]
\end{equation}
\begin{equation}\label{noi_grad}
    \frac{\partial \mathcal{L}_\mathcal{N}}{\partial\theta}=\sum_{i}^N\bigg[\frac{\partial (\mathcal{H}^i_b\circ f^i_\theta)}{\partial f^i_\theta}\frac{\partial f^i_\theta}{\partial \theta}+\frac{\partial (\mathcal{H}_{b,d}\circ f^i_\theta)}{\partial f^i_\theta}\frac{\partial f^i_\theta}{\partial \theta}\bigg]
\end{equation}
}

The key to \eqnref{sig_grad} and \eqnref{noi_grad} is the partial derivative of the birth and death functionals with respect to $f_\theta$, which can be determined by their cofaces. Intuitively, for a topological feature, its partial derivative is calculated by fetching the indices of birth and death stored in the corresponding coface.
Denote by $\varsigma$ the cofaces corresponding to all topological features $\{(b_i, d_i)\}$. Since $\varsigma$ itself is essentially a set composed of points, line segments, squares, and cubes, it can be structurally expressed as a tensor-like format:
$
    \varsigma := T^{0,1,2,3}_{1,2,...,n}[\mathbf{f}],
$
where $\mathbf{f}$ is a two-dimensional vector basis, which signifies the indices of $(b_i,d_i)$ in a specific region. Corresponding the significant or noise features, $n=S\;\text{or}\;N$. We denote by $(\mathbf{i}_b, \mathbf{i}_d)$ the index tuple of all topological features, where $\mathbf{i}_b$ and $\mathbf{i}_d$ indicate respectively the indices of birth and death features. Obviously, \eqnref{sig_grad} and \eqnref{noi_grad} can be rewritten as 
$\frac{\partial \mathcal{L}_\mathcal{S}}{\partial\theta}=\frac{\partial \mathcal{L}_\mathcal{S}}{\partial f_\theta}\frac{\partial f_\theta}{\partial\theta}$ and $\frac{\partial \mathcal{L}_\mathcal{N}}{\partial\theta}=\frac{\partial\mathcal{L}_\mathcal{N}}{\partial f_\theta}\frac{\partial f_\theta}{\partial\theta}$.
With this in hand, we have the following:
{
\small
\begin{equation}
    \frac{\partial \mathcal{L}_\mathcal{S}}{\partial f^i_\theta}=\begin{cases}
        a & \text{if $i\in\mathbf{i}_b$ and $l>0$}\\
        -a & \text{if $i\in\mathbf{i}_d$} \\
        0 & \text{Otherwise}
    \end{cases}
\end{equation}
}where $a$ is the gradient value from the first connectivity loss component, and $l\in\{0,1,2,3\}$. Similarly, we have the following relationship for the noise features:
{
\small
\begin{equation}
    \frac{\partial\mathcal{L}_\mathcal{N}}{\partial f_\theta}=\begin{cases}
        \vartheta & \text{if $i\in\mathbf{i}_b$} \\
        \vartheta & \text{if $i\in\mathbf{i}_d$ and $l>0$}\\
        0 & \text{Otherwise}
    \end{cases}
\end{equation}
}where $\vartheta$ is the gradient value from the second connectivity loss component.
Therefore, during the backpropagation, the differentiability is ensured such that 
{
\small
\begin{equation}
    \frac{\partial \mathcal{L}}{\partial\theta}=\frac{\partial \mathcal{L}_g}{\partial\theta}+\lambda_1\frac{\partial \mathcal{L}_\mathcal{S}}{\partial\theta} + \lambda_2\frac{\partial \mathcal{L}_\mathcal{N}}{\partial\theta}.
\end{equation}
}

\section{Theoretical Analysis}
\label{sec:analysis}
According to the unified loss in the above section, we present the theoretical analysis in this section to show the explicit convergence to the first-order critical point by using the stochastic gradient descent algorithm. Although previous works~\citep{carriere2021optimizing,scoccola2024differentiability} have considered connectivity loss in their optimization, subgradient descent was still used for the updates. The adoption of subgradient may cause divergence as it can be randomly sampled from a Clarke subdifferential at a point. As in our study, we connect the parametric model with the persistence through a key concept of coface so as to address the non-differentiability issue. 
Given the optimal solution, we will then show it theoretically provides the guarantee of connectivity for the topological surface reconstruction. We defer some analysis and detailed proof to Appendix.

\subsection{Stochastic gradient descent}
$\mathcal{L}_c$ is definable since it is differentiable. In addition, $\mathcal{L}_{g}$ is parameterized by $\theta\in\mathbb{R}^d$ and typically assumed to be continuously differentiable in deep learning-based stochastic optimization~\citep{shah2024leveraging}, leading it to be definable. As a consequence, $\mathcal{L}$ is also \textit{differentiable}. To minimize $\mathcal{L}$, the stochastic gradient descent (SGD)~\citep{bottou2018optimization} algorithm is employed accordingly, which has the update law  as follows:
{
\small
\begin{equation}\label{subgradient_update}
    \theta_{t+1}=\theta_t-\alpha_t(\nabla\mathcal{L}(\theta_t)+\zeta_t),
\end{equation}
}where $\alpha_t$ is the learning rate satisfying assumptions imposed below, $\nabla\mathcal{L}(\theta_t)$ is the gradient, and $\zeta_t$ is a sequence of random variables that signify the stochastic noise caused by calculating the gradient $\nabla\mathcal{L}(\theta_t)$. Specifically, the following technical assumptions hold true throughout the analysis.
\begin{assumption}\label{assump_1}
(a) For any $t, \alpha_t\geq 0$, $\sum_{t=1}^\infty=+\infty$, and $\sum_{t=1}^\infty\alpha^2_t<+\infty$;
(b) $\textnormal{sup}\|\theta_t\|<+\infty$ almost surely;
(c) Denoting by $\mathcal{F}_t$ the increasing sequence of $\sigma$-algebras $\mathcal{F}_t=\sigma(\theta_\iota,\nabla\mathcal{L}(\theta_\iota),\zeta_\iota,\iota<t)$, there exists a function $e:\mathbb{R}^d\to\mathbb{R}$, which is bounded on compact sets such that almost surely, for any $t$,
            $\mathbb{E}[\zeta_t|\mathcal{F}_t]=0\;\textnormal{and}\;\mathbb{E}[\|\zeta_t\|^2|\mathcal{F}_t]<e(\theta_t).$
(d) There exists $\mathcal{L}^*=\textnormal{min}_{\theta\in\mathbb{R}^{d}}\mathcal{L}(\theta)$ such that $\mathcal{L}^*>-\infty$.
\end{assumption}
\assmref{assump_1} (a) indicates that the learning rate is not summable but square-summable, which can easily be satisfied if we take $\alpha_t=1/t$. \assmref{assump_1} (b) is typically easy to verify for most functions. For \assmref{assump_1} (c), it is a fairly standard one that has widely been imposed in numerous works~\citep{bottou2010large,bottou2018optimization}. It suggests that $\zeta_t$ has zero mean and bounded second moment, conditioned on the past. This can practically be achieved by taking a sequence of independent and centered variables with bounded variance, such as a Gaussian distribution.
\begin{theo}\label{theorem_1} Suppose that \assmref{assump_1} holds and that $\mathcal{L}_{g}$ is continuously differentiable.
    Let $\mathcal{C}_K$ be a cubical complex and $\Phi:\mathbb{R}^K\to\mathbb{R}^{|\mathcal{C}_K|}$ a family of filtrations of $\mathcal{C}_K$ that is definable in an $o$-minimal structure. A persistence map $\mathcal{M}:\mathcal{A}_{|\mathcal{C}_K|}\to\mathbb{R}^{|\mathcal{C}_K|}$ is defined to indicate the PD assigned to each filtration of $\mathcal{C}_K$. Let $F:\mathbb{R}^{\mathcal{C}_K}\to\mathbb{R}$ be a definable function of persistence such that $\mathcal{L}_c=F\circ\mathcal{M}\circ\Phi$ is locally Lipschitz. Thus, the limit points of the sequence $\{\theta_t\}$ generated by \eqnref{subgradient_update} are critical points of $\mathcal{L}$ and the sequence $\mathcal{L}(\theta_t)$ converges to $\mathcal{L}^*$ almost surely.
\end{theo}
Note that $\mathcal{C}_K$, the $o$-minimal structure and $\mathcal{A}_{|\mathcal{C}_K|}$ are all defined in the Appendix due to the space limit. \theoremref{theorem_1} theoretically guarantees the convergence of $\mathcal{L}$ by using the SGD algorithm, which essentially attains the optimal parameter $\theta^*$ and topological features that satisfy
$
    \varepsilon^*=\{(b^*_i,d^*_i)|b^*_i=b^*_{i+1}, d^*_i=d^*_{i+1}, b^*_i\neq d^*_i\}, \forall i\in[|\mathcal{C}_K|]
$ in this study. 

This result bridges the theoretical gap in a recent work~\cite {dong2022topology}, where they presented a similar problem formulation but did not show theoretical convergence. A similar conclusion was also reached in \citet{carriere2021optimizing}. Though both works have used the same first-order optimization algorithm, the significant distinction is the loss function. In their case, they have only considered the optimization for persistent homology without any combination with parameterized models, which only consider the connectivity. However, in our study, we focus primarily on the unified loss, which is more complex.

\subsection{Theoretical guarantee for connectivity}
Given the optimal solution $\theta^*$ obtained by optimizing the loss $\mathcal{L}$, we obtain a parameterization for SDF such that ideally for each topological feature, $b_i^*\neq d^*_i$. However, as the unified loss involves two competing objectives, it is numerically difficult to get $(b^*_i,d^*_i)$ satisfying condition in $\varepsilon^*$. If we define $\xi_i=|b_i-d_i|$ based on the definition of persistence of the topological feature, then $\xi_i$ typically lies in some interval to ease the optimization in practice. In this work, we study the connectivity in terms of density behavior and separatedness. 
The definition for a connected set is introduced in the sequel.
\begin{definition}($\alpha\sim\beta$-connected set\citep{hofer2019connectivity})\label{defi_4} Let $D\subset\mathbb{R}^3$ be a finite set and let $\{\xi_i\}_{i=1}^{|\mathcal{C}_K|}$ be the increasing sequence of pairwise distance values of $D$. Hence, $D$ is $\alpha\sim\beta$-connected if and only if:
$\alpha=\textnormal{min}_{i\in[|\mathcal{C}_K|]}\xi_i\;\textnormal{and}\:\beta=\textnormal{max}_{i\in[|\mathcal{C}_K|]}\xi_i$.
\end{definition}
Based on the above definition, it implies that during training $\mathcal{L}$ controls properties of $\mathbf{Q}$ explicitly, and at the convergence, $\mathbf{P}$ is $\alpha\sim\beta$-connected. It should be noted that the optimization directly applies to the input space instead of the latent space, as done in~\citep{hofer2019connectivity}, which stands as one significant difference. We next investigate the impact of $\alpha\sim\beta$-connectedness on the density around a feature representation by introducing the following theorem. Some relevant definitions, including $m\sim\epsilon$-dense set and $\epsilon$-separated, are defined in Appendix.
\begin{theo}\label{theorem_2}
    Let $2\leq k \leq m$ and $M\subset\mathbb{R}^3$ with $|M|=m$ such that for each $D\subset M$ with $|D|=k$, it holds that $D$ is $\alpha\sim\beta$-connected. Then $M$ is $(m-k+1)\sim\beta$-dense.
\end{theo}
\theoremref{theorem_2} quantifies the number of points in the neighborhood of each point in $M$, within distance $\beta$, which is $m-k+1$. This result facilitates the theoretical understanding of the impact of the given point cloud $\mathbf{P}$ on connected components in the topological surface reconstruction in this study. Taking another different perspective of connectivity, we can study the separation of points in $M$ as well. Intuitively speaking, when $m$ increases, the separation of points in $M$ is expected to decrease since it becomes denser. A formal result in \theoremref{theorem_3} quantifies this in the Appendix. 

We also remark on the difference between our results and those in~\citet{hofer2019connectivity}. Despite the fact that some proof techniques are adapted from the latter, we emphasize that our conclusions have been generally dedicated to a different application scenario with topology-controllable neural implicit representation. Also, in their work, they do not include a detailed analysis for the optimization of the parameterized model, with the assumption that reconstruction loss jointly with the connectivity loss is minimized.

\section{Experiments}
\label{experiments}

We perform an extensive evaluation of our method for the task of surface reconstruction. Please see the Supplement for additional results and ablation studies. 

We compare our method against different surface reconstruction methods---Poisson Surface Reconstruction (PSR)~\citep{kazhdan2006poisson}, Neural-Pull (NP)~\citep{ma2020neural}, Implicit Geometric Regularization (IGR)~\citep{gropp2020implicit}, Divergence-guided shape implicit representation for unoriented point clouds (DiGS)~\citep{ben2021digs}, and Neural Singular Hessian (NSH)~\citep{wang2023neural}. Each of these methods (except for PSR) is able to work with unoriented point clouds.

We evaluate our method on several different surface reconstruction datasets. We start with the surface reconstruction benchmark (SRB)~\citep{berger2013benchmark} and utilize the variant made widely available by \citet{williams2019deep} that includes ground truth meshes and corresponding point clouds. SRB consists of five different shapes---Anchor, Daratech, Dc, Gargoyle, and Lord Quas, each with varying geometric features. We also use five different shapes from the DFAUST~\citep{bogo2017dynamic} dataset, which consists of $\approx$40k human scans captured at various time points

To better showcase the versatility of our approach, We also test on eight different plant geometries, each with a unique set of geometric features---particularly thin leaves and narrow connection between the leaves and the stalk. In addition, we also test on four challenging hollow and thin models---\textit{Torus}, \textit{Hollow Ball}, \textit{Eiffel Tower}, and \textit{Lamp}. To compare the reconstruction accuracy, we compute the mesh of the object from the SDF evaluated at a resolution of $256^3$ using Marching Cubes~\citep{lorensen1998marching}. We then sample points from the reconstruction to compute the one-sided Chamfer distance (see Supplement for the equations). Furthermore, we use the final predicted SDF for each method to calculate the topological loss.

\begin{figure}[!b]
    \centering
    \includegraphics[width=0.99\linewidth,trim={0.1in 0.7in 0.1in 0.7in},clip]{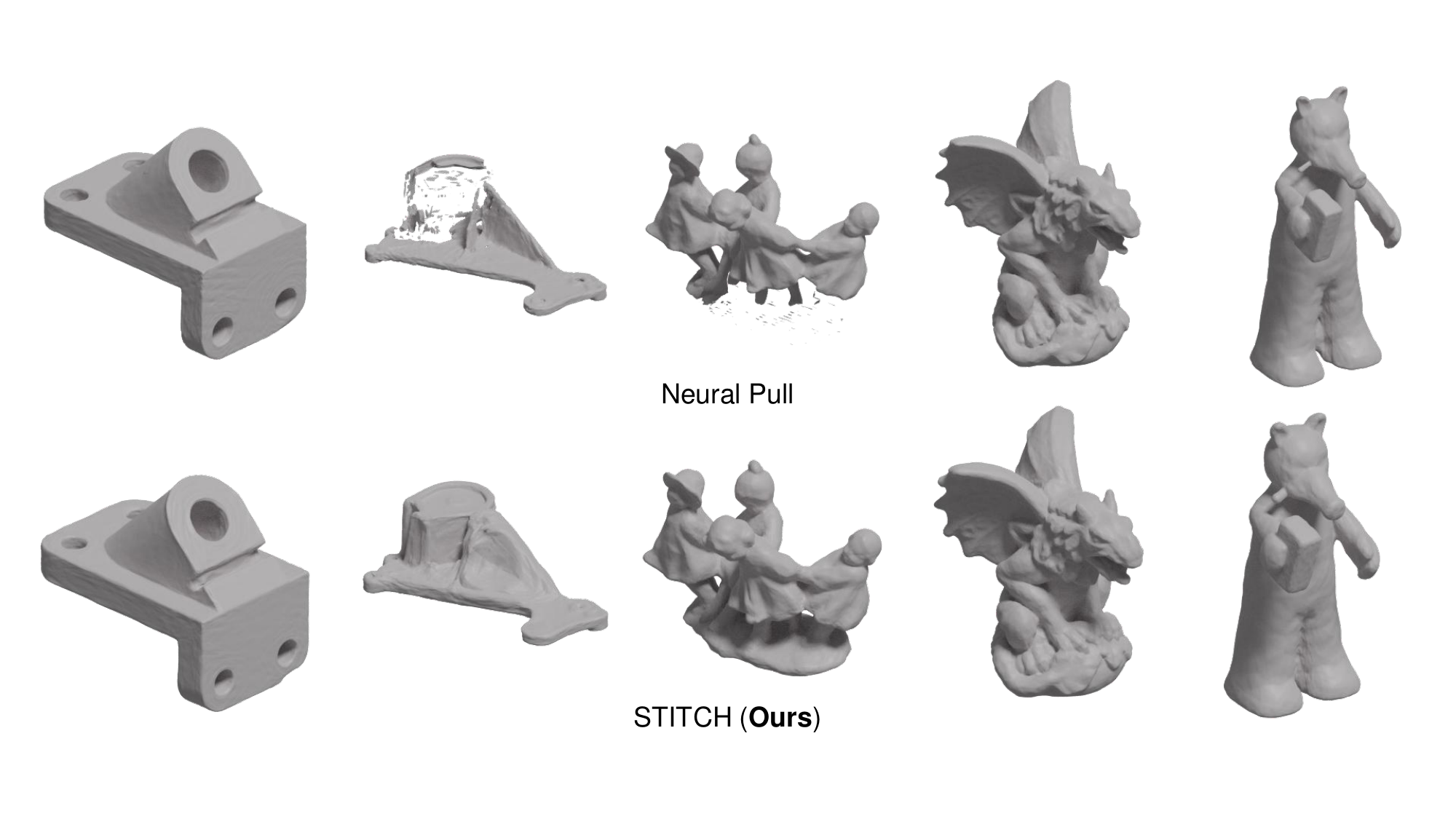}
    \caption{Comparison of the reconstructions of the SRB dataset by our proposed method with Neural Pull.}
    \label{fig:srb_result}
\end{figure}

\begin{table}[t!]
\centering
\caption{Comparison of one-sided CD (reconstruction to ground truth scan) for models from the SRB.}
\label{tab:srb_one_sided_pred_gt}
\scriptsize
\setlength{\tabcolsep}{3pt}  
\begin{tabular}{@{}l@{\hspace{4pt}}cccccccc@{}} 
\toprule
Model & PSR & NP & IGR & DiGS & OG-INR & NSH & STITCH (Ours) \\
\midrule
Anchor    & \textbf{0.0076} & 0.0097 &   -    & 0.0104 & 0.0095 & 0.0079 & 0.0095\\
Daratech  & \textbf{0.0057} & 0.0071 &   -    & 0.0073 & 0.0061 & 0.0060 & 0.0089\\
Dc        & \textbf{0.0061} & 0.0080 & 0.2294 & 0.0080 & 0.0063 & 0.0062 & 0.0081\\
Gargoyle  & 0.0078 & 0.0099 &   -    & 0.0094 & 0.0082 & \textbf{0.0073} & 0.0101\\
Lord\,Quas & \textbf{0.0046} & 0.0062 & 0.0852 & 0.0063 & 0.0048 & 0.0047 & 0.0063\\
\midrule
Mean       & \textbf{0.0064} & 0.0082 & -      & 0.0083 & 0.0070 & \textbf{0.0064} & 0.0086 \\
Std.\,Dev.    & 0.0012 & 0.0014 & -      & 0.0015 & 0.0017 & \textbf{0.0011} & 0.0013 \\
\bottomrule
\end{tabular}
\end{table}

\begin{table}[t!]
\centering
\caption{Comparison of significant features topological loss term for models from SRB.}
\label{tab:srb_tda_metrics}
\scriptsize
\setlength{\tabcolsep}{3pt}
\begin{tabular}{@{}l@{\hspace{4pt}}ccccc@{}} 
\toprule
Model & NP & DiGS & OG-INR & STITCH (Ours) \\
\midrule
Anchor & 2.7066 & 5.4486 & 18.3466 & \textbf{0.3280}\\
Daratech & 1.1846 & 5.9938 & 9.2916 & \textbf{0.4323}\\
Dc & 3.8121 & 9.1217 & 20.1540 & \textbf{1.5936}\\
Gargoyle & 3.3629 & 13.6594 & 12.9182 & \textbf{0.7460}\\
Lord\,Quas & 2.6336 & 5.2952 & 13.7763 & \textbf{0.1930}\\
\bottomrule
\end{tabular}
\end{table}

\begin{figure}[!t]
    \centering
    \includegraphics[width=0.99\linewidth,trim={1.5in 1.2in 1.5in 1.2in},clip]{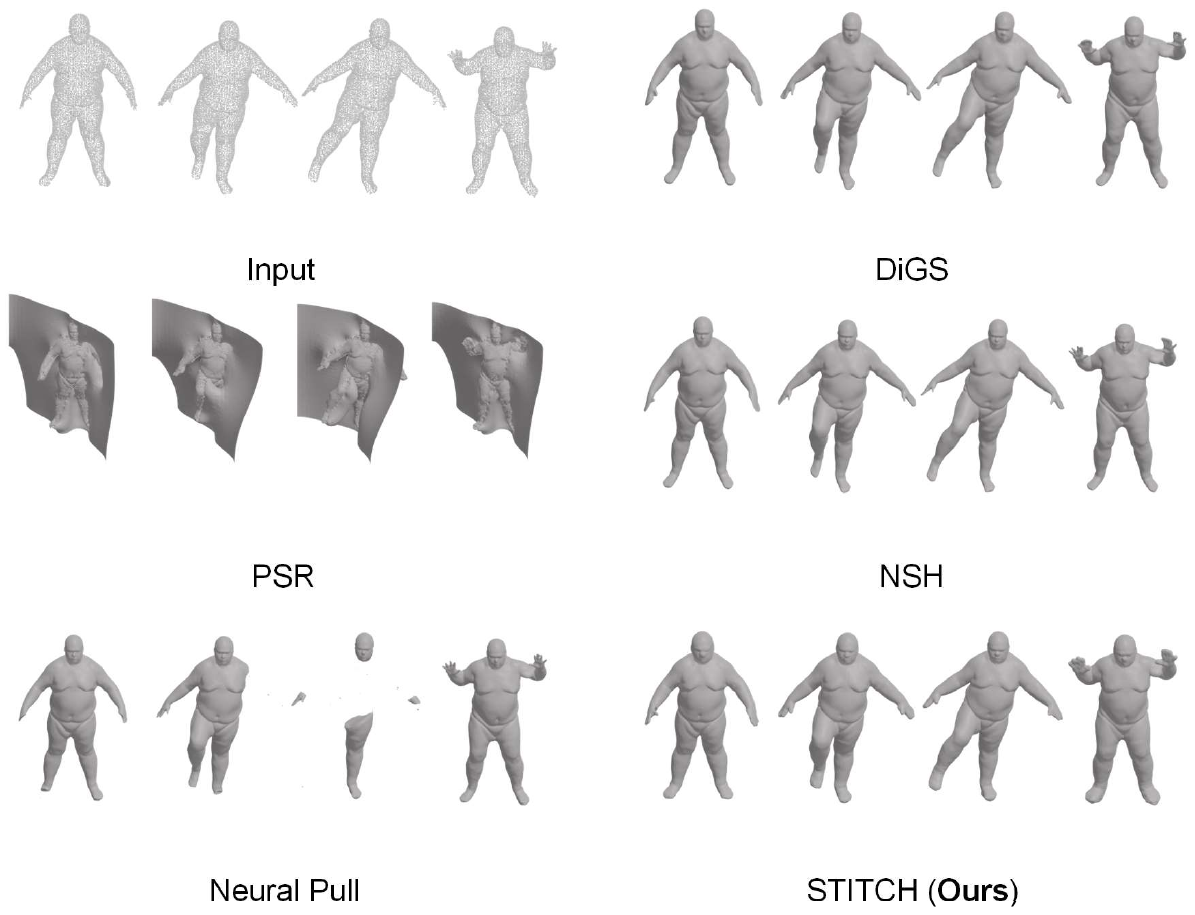}
    \caption{Comparison of different methods for the reconstruction of four models from the DFAUST dataset.}
    \label{fig:dfaust_result}
\end{figure}

\begin{table}[t!]
\centering
\caption{Comparison of one-sided CD (reconstruction to ground truth registration) for five shapes from the DFAUST dataset.}
\label{tab:dfaust_one_sided_pred_gt}
\scriptsize
\setlength{\tabcolsep}{3pt}  
\begin{tabular}{@{}l@{\hspace{4pt}}cccccccc@{}} 
\toprule
Model & PSR & NP & IGR & DiGS & OG-INR & NSH & STITCH (Ours) \\
\midrule
Shape 1 & 0.2618 & 0.0081 & 0.0179 & 0.0046 & 0.0042 & \textbf{0.0040} & 0.0084\\
Shape 2 & 0.2911 & 0.0179 & 0.0275 & 0.0044 & 0.0044 & \textbf{0.0043} & 0.0076\\
Shape 3 & 0.2538 & \textbf{0.0045} & 0.0095 & 0.0048 & 0.0050 & 0.0059 & 0.0086\\
Shape 4 & 0.1197 &   -    & 0.0115 & 0.0062 & \textbf{0.0039} & 0.0040 & 0.0076\\
Shape 5 & 0.2911 & 0.0054 & 0.0080 & 0.0045 & 0.0044 & \textbf{0.0041} & 0.0078\\
\midrule
Mean    & 0.2435 &    -   & 0.0149 & 0.0049 & \textbf{0.0044} & 0.0045 & 0.0080 \\
Std.\,Dev. & 0.0637 &    -   & 0.0072 & 0.0007 & \textbf{0.0004} & 0.0007 & \textbf{0.0004}\\
\bottomrule
\end{tabular}
\end{table}

\begin{figure*}[!t]
    \centering
    \includegraphics[width=0.99\linewidth,trim={0.3in 3.9in 0.3in 0.0in},clip]{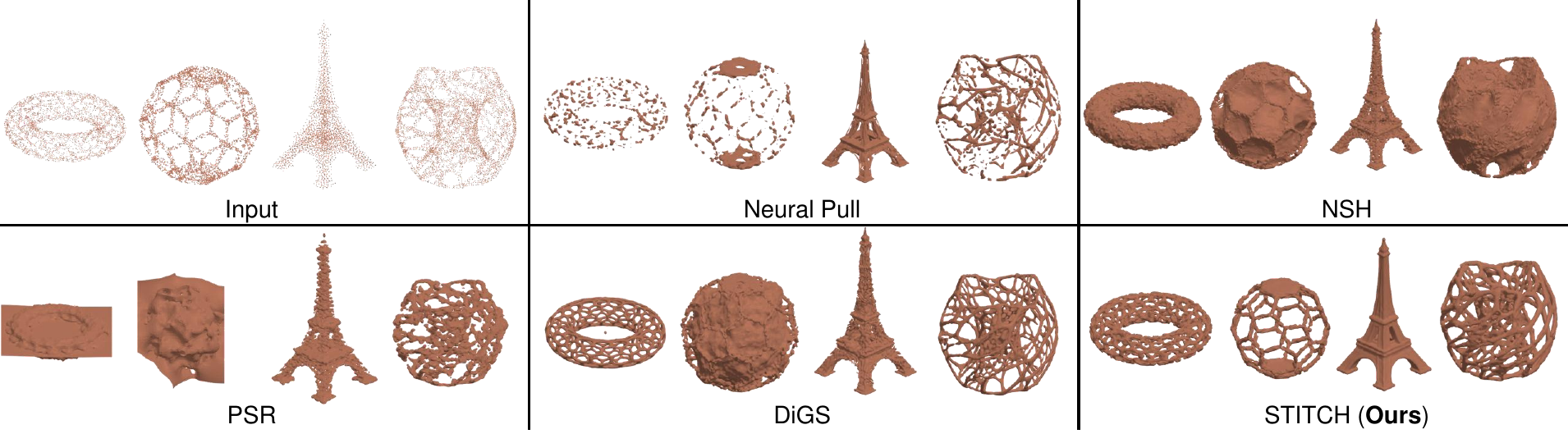}
    \caption{Comparison of different reconstruction methods with challenging point cloud data with thin structures.}
    \label{fig:shapes_result}
\end{figure*}

\begin{figure}[!t]
    \centering
    \includegraphics[width=0.99\linewidth,trim={1.6in 2.4in 1.6in 0.0in},clip]{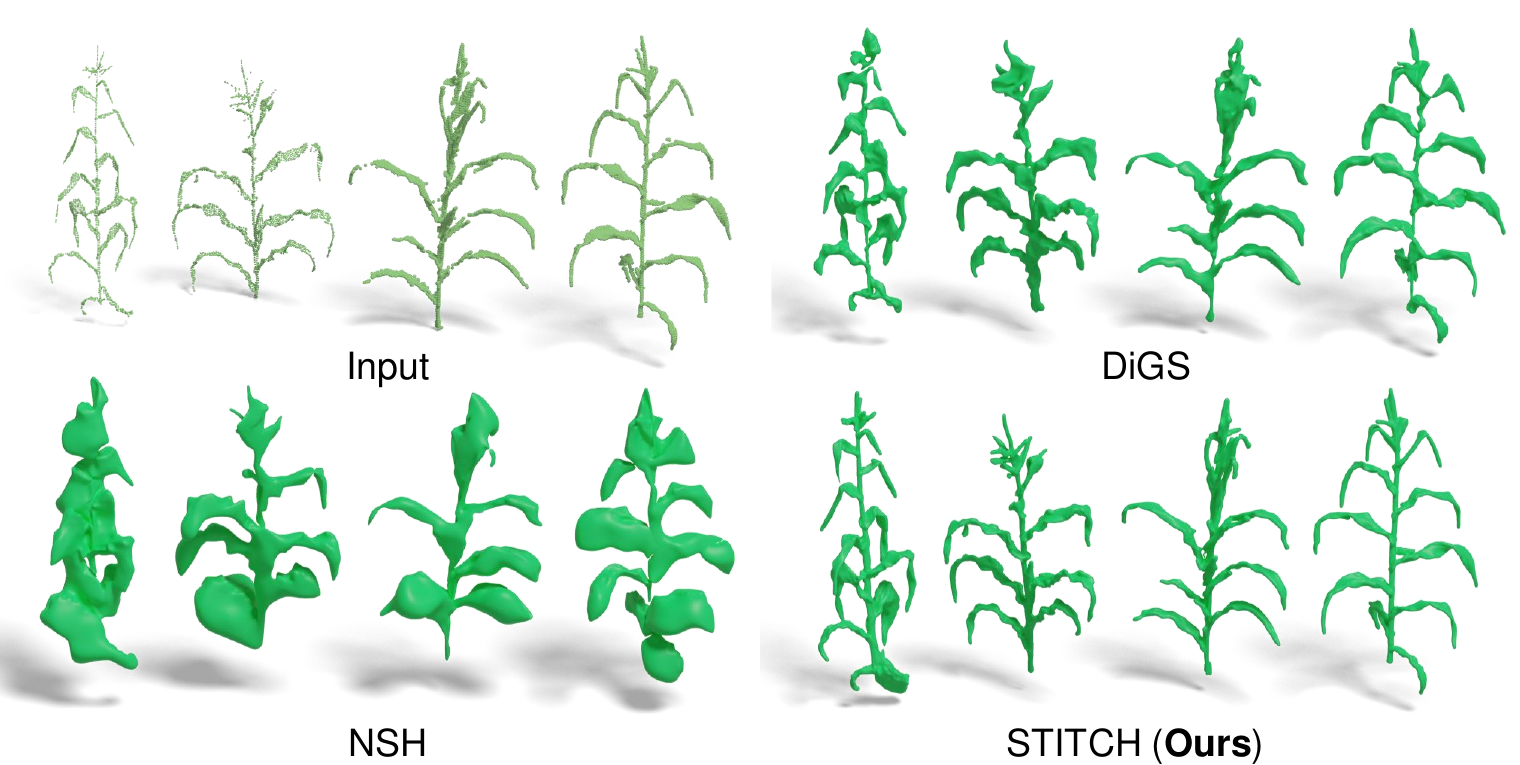}
    \caption{Comparison of different methods for surface reconstruction of plants.}
    \label{fig:plant_result}
\end{figure}

\begin{table}[t!]
\centering
\caption{Comparison of one-sided CD (reconstruction to ground truth) for eight different plants.}
\label{tab:plants_one_sided_pred_gt}
\scriptsize
\setlength{\tabcolsep}{3pt}  
\begin{tabular}{@{}l@{\hspace{4pt}}cccccccc@{}} 
\toprule
Model & PSR & NP & IGR & DiGS & OG-INR & NSH & STITCH (Ours) \\
\midrule
Plant 1 & 0.0755 & - & 0.5161 & \textbf{0.0110} & - & 0.0225 & 0.0132\\
Plant 2 & 0.2477 & 0.0635 & - & 0.0132 & - & 0.0306 & \textbf{0.0067}\\
Plant 3 & 0.2669 & - & - & 0.0096 & 0.0069 & 0.0271 & \textbf{0.0067}\\
Plant 4 & 0.2004 & - & - & 0.0112 & 0.0067 & 0.0281 & \textbf{0.0066}\\
Plant 5 & 0.2422 & - & 0.3594 & 0.0112 & 0.0070 & 0.0491 & \textbf{0.0065}\\
Plant 6 & 0.1590 & - & - & 0.0127 & 0.0090 & 0.0229 & \textbf{0.0073}\\
Plant 7 & 0.1214 & - & - & 0.0170 & \textbf{0.0080} & 0.0403 & 0.0115\\
Plant 8 & 0.3388 & - & 0.4105 & 0.0109 & - & 0.0373 & \textbf{0.0069}\\
\midrule
Mean    & 0.2065 & - & - & 0.0121 & - & 0.0322 & \textbf{0.0082}\\
Std.\,Dev. & 0.0797 & - & - & \textbf{0.0021} & - & 0.0087 & 0.0025\\
\bottomrule
\end{tabular}
\end{table}

The evaluated metrics for the SRB dataset are shown in \tabref{tab:srb_one_sided_pred_gt} and \tabref{tab:srb_tda_metrics}. STITCH performs competitively on the SRB dataset with a low standard deviation and is able to maintain stable performance close to top-performing baselines such as OG-INR and NSH. While PSR has slightly better results on some shapes, STITCH maintains consistently good alignment and is particularly effective in more challenging shapes (\figref{fig:srb_result}). Topologically, STITCH outperforms all other methods with significantly lower significant feature loss across all tested shapes compared to other methods. This indicates that STITCH is able to preserve relevant geometric features and produce a single connected component. On the DFAUST dataset (\tabref{tab:dfaust_one_sided_pred_gt}), STITCH is able to achieve competitive mean and standard deviation values and its performance is in line with DIGS and OG-INR.

\figref{fig:shapes_result} shows a comparison of different reconstruction methods on 3D shapes with thin, challenging structures that have a wide variation in the point cloud density and noise. This comparison highlights the strengths of our method in handling thin and intricate shapes. Unlike other methods that either oversmooth or produce artifacts, STITCH achieves detailed reconstructions that closely match the input shapes. 

We showcase a similar comparison for plant geometries in \figref{fig:plant_result} and their metrics in \tabref{tab:plants_one_sided_pred_gt}. Plant leaves are often prone to occlusion (given the proximity of neighboring leaves), leading to inaccurate reconstructions or blob-like surfaces. STITCH outperforms baseline methods and effectively captures the relevant features pertaining to each plant geometry, and is able to reconstruct the surfaces with minimal blob-like surfaces. We attribute the tiny amount of blobs to the bias in our loss for a single connected component.

\section{Conclusions and Broader Impacts}
\label{sec:conclusions}

We have presented a general differentiable approach for incorporating topological loss terms with implicit surface reconstruction. Our approach produces a single connected component of the reconstructed surface, especially with sparse point cloud data. Further, we prove that our approach converges and is guaranteed to result in a single connected component. The primary use case of such a reconstruction is enabling the direct use of reconstructed objects in physically-based simulations (such as flow or structure simulations), which usually require well-defined inside and outside domains enclosed by a 2-manifold surface. 

\noindent\textbf{Limitations:} There is a possibility of edge cases where certain subtle topological features might be incorrectly penalized due to short persistence. The loss term we use in this work encourages a single connected surface for the reconstructed model, which might incorrectly combine 2 separate objects. Furthermore, our current approach is limited to 0-dimensional topological features, owing to the computational complexity of computing higher-dimensional features. 

Our method is quite generic in integrating TDA-based losses for geometric deep-learning and may impact applications such as computer graphics, physics simulations, and engineering design. Care should be taken to ensure that these applications are deployed responsibly. Future works include exploring the incorporation of higher-order topological features and extending the reconstruction to topological features beyond a single connected component.

{
    \small
    \bibliographystyle{ieeenat_fullname}
    \bibliography{main}
}

\clearpage
\setcounter{page}{1}
\maketitlesupplementary
\appendix

\section{Surface Reconstruction Methods}
Traditional methods such as Voronoi diagrams~\citep{amenta1998new} and Delaunay triangulations~\citep{boissonnat1984geometric,kolluri2004spectral} were initially developed for generating a triangular mesh from points; however, these methods are not robust to noise. Implicit methods handle noise in the data much better than pure geometric approaches. For example, the most common approach---Poisson surface reconstruction~\citep{kazhdan2006poisson}---finds a global indicator function based on the estimate of a local tangent plane~\citep{hoppe1992surface}. \citet{carr2001reconstruction} used radial basis functions at the center of the point coordinates to compute an implicit function. The level-set method has also been used to fit an implicit surface to a point cloud without normals~\citep{zhao2001fast}. 

Leveraging the generalized function approximation nature of neural networks has recently allowed the replacement of the implicit function with a neural network under the class of \emph{Implicit Neural Representations}~\citep{atzmon2020sal,atzmon2021sald, liu2022learning,ben2021digs}. Specifically, if the implicit function is the distance field, it allows for the use of physics-informed methods that solve the Eikonal equation using a boundary value problem for surface reconstruction~\citep{sitzmann2020implicit,gropp2020implicit}. This approach guides the distance function to have a value of zero on the underlying surface represented by the point cloud. Neural occupancy functions have also been used for surface reconstruction~\citep{mescheder2019occupancy}. By extension, implicit representations have also been used for representing 3D scenes~\citep{mildenhall2020nerf,yu2021pixelnerf,wang2021neus,muller2022instant}.

Often, mesh extraction from these approaches is done via the well-known marching-cubes algorithm~\citep{lorensen1998marching}, and meshes for point clouds with thin and sharp features are not guaranteed to be manifold meshes. None of the conventional neural implicit approaches tackle maintaining topological accuracy while capturing thin and sharp features commonly found in plants, chairs, and the human body.

\section{Visual Illustration of Cubical complex and Persistent Homology}
\label{appsec:vis}

\begin{figure*}[!t]
    \centering
    \includegraphics[width=0.88\linewidth,trim={0.0in 2.25in 0.0in 1.75in},clip]{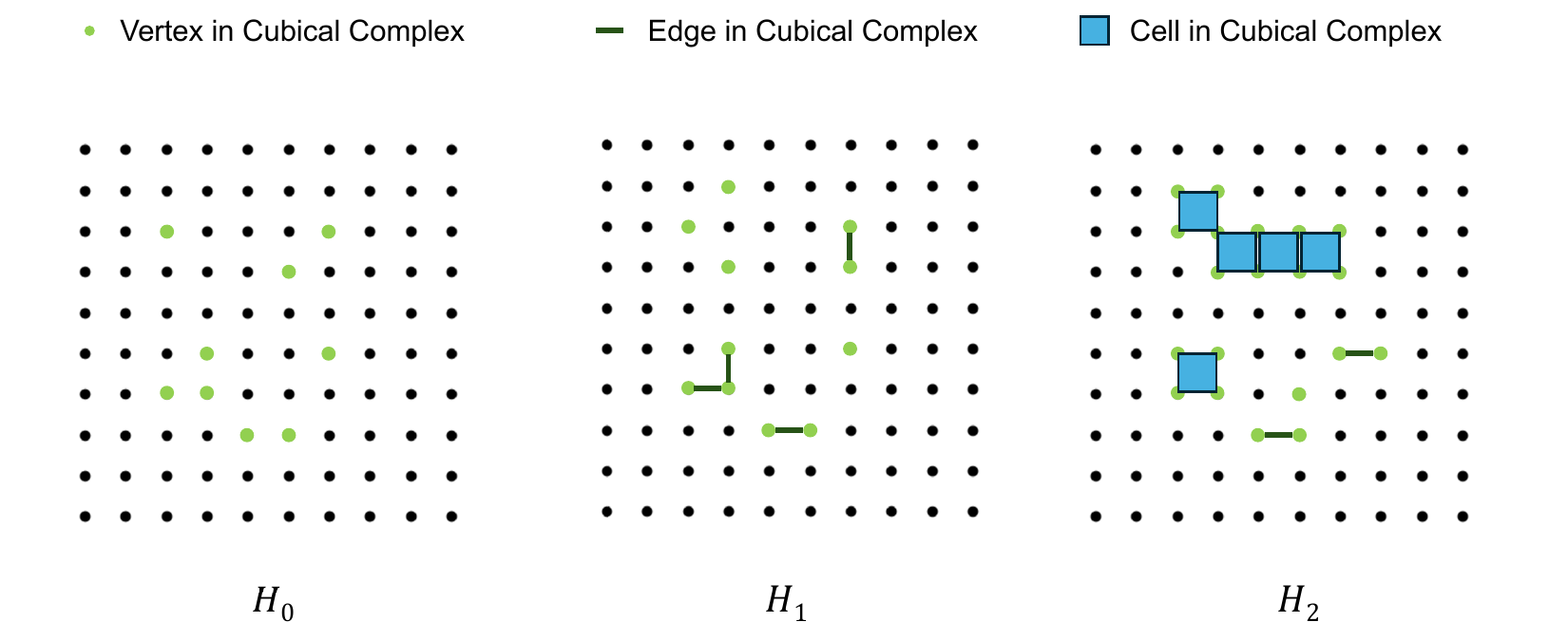}
    \caption{An example of a 2D cubical complex depicting multiple topological features at increasing thresholds. From left to right, as the threshold increases, higher dimensional features like edges and cells appear. $H_0$, $H_1$, and $H_2$ correspond to 0, 1, and 2-dimensional features, respectively.}
    \label{fig:topological_features}
\end{figure*}

\begin{figure*}[!t]
    \centering
    \includegraphics[width=0.85\linewidth,trim={0.0in 2.25in 0.0in 2.125in},clip]{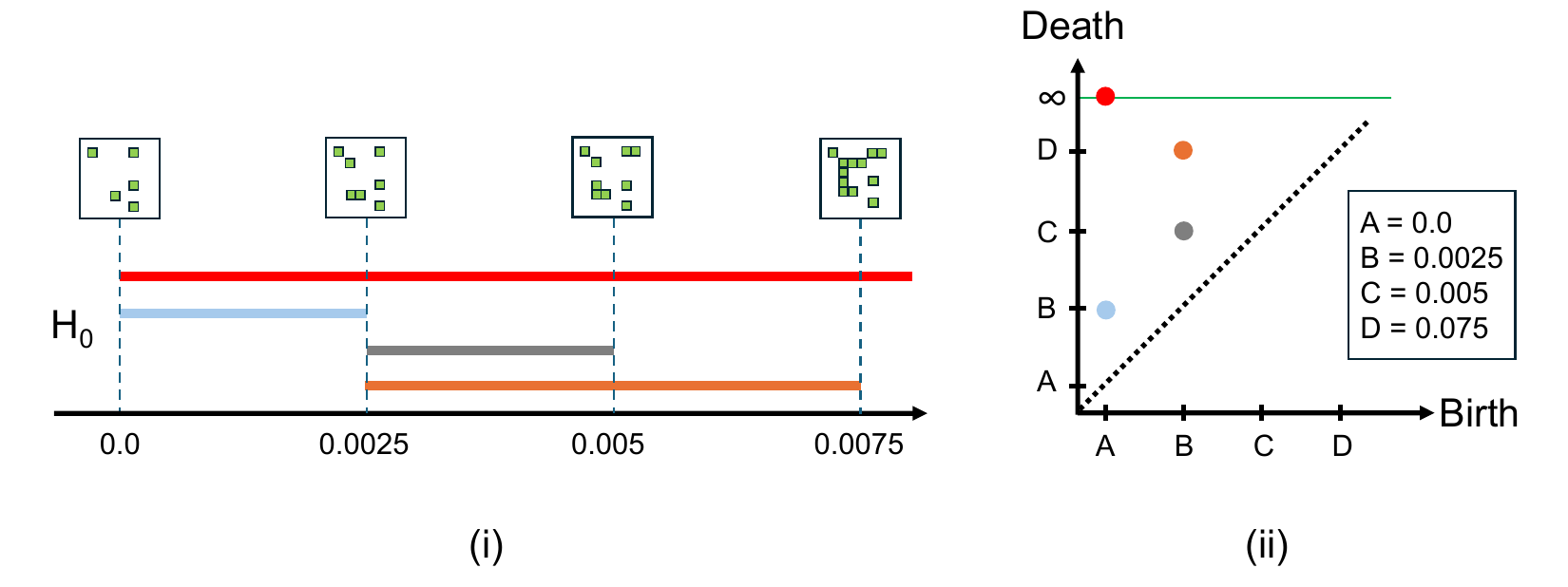}
    \caption{An example of the filtration process, capturing various features at increasing thresholds. We only show appearance of 0-dimensional features is showcased. In (i), we show an example of a persistence barcodes and are generated by increasing the threshold, making various 0-dimensional features appear and disappear. Each colored horizontal bar represents a feature and its starting and ending threshold correspond to birth and death values, respectively. In (ii), we show the corresponding persistent diagram (PD). Each barcode in (i) is a point on the PD Similar colors are used in (i) and (ii) for visual clarity.}
    \label{fig:filtration_example}
\end{figure*}

\figref{fig:topological_features} shows various topological features of a cubical complex for a 2D example. \figref{fig:filtration_example} shows an example of the filtration process for a 2D cubical complex.

\section{Additional Analyses and Proofs}
In this section, we include additional preliminaries, analysis and proof that characterize our method. We present the theorem statements for the completeness.

\subsection{Additional preliminaries and analysis}
We define the maximum of all points in a cubical complex as follows to characterize the filtration function.
\begin{definition}\label{defi_1}
Let $\mathcal{S}$ be the underlying space of a cubical complex $\mathcal{C}$, denoted by $\mathcal{C}(\mathcal{S})$, and $h:\mathcal{S}\to\mathbb{R}$ be a real-valued function such that the value of a cube $\tau$ in $\mathcal{C}(\mathcal{S})$ is defined as the maximum over all points in the cube, i.e., $h(\tau)=\textnormal{max}_{p\in\tau}h(p)$.
\end{definition}
Practically speaking, $h(\tau)$ can be estimated by using the maximum of all the vertices of a cube. With this in hand, we are now ready to state the sublevel filtration function.

\begin{definition}\label{defi_2}
    Denote by $\mathcal{C}_K=\{\tau\in\mathcal{C}(\mathcal{S})|h(\tau)\leq K, K\in\mathbb{R}\}$ a cubical complex . The sublevel filtration of $\mathcal{C}(\mathcal{S})$, $\Phi(h):\mathbb{R}^K\to\mathbb{R}^{|\mathcal{C}_K|}$, filtered by $h$ is a nested sequence of filtered cubical complexes, which satisfies $\mathcal{C}_{K_i}\subseteq\mathcal{C}_{K_j}$ for all $K_i\leq K_j, K_i, K_j\in\mathbb{R}$.
\end{definition}

We introduce a concept called o-minimal structure~\citep{fujita2023locally,guerrero2023p}, which can be defined as follows.
\begin{definition}(o-minimal structure~\citep{carriere2021optimizing})\label{defi_3}
    An o-minimal structure on the filed of real numbers $\mathbb{R}$ is a collection $(B_n)_{n\in\mathbb{N}}$, where $B_n$ is a set of subsets of $\mathbb{R}^n$ such that:
    \begin{itemize}
        \item $B_1$ is exactly the collection of finite unions of points and intervals;
        \item all algebraic subsets of $\mathbb{R}^n$ are in $B_n$;
        \item $B_n$ is a Boolean subalgebra of $\mathbb{R}^n$ for any $n\in\mathbb{N}$;
        \item if $A\in B_n$ and $C\in B_r$, then $A\times C\in B_{n+r}$;
        \item if $\pi:\mathbb{R}^{n+1}\to\mathbb{R}^n$ is the linear projection onto the first $n$ coordinates and $A\in B_n$, then $\pi(A)\in B_{n+1}$.
    \end{itemize}
\end{definition}

\begin{lemma}\label{lemma_1}
    Let $\mathcal{C}_K$ be a cubical complex and $\Phi:\mathbb{R}^K\to\mathbb{R}^{|\mathcal{C}_K|}$ is a definable filtration. Denote by $\mathcal{A}_{|\mathcal{C}_K|}$ the set of vectors in $\mathbb{R}^{|\mathcal{C}_K|}$. The map $\mathcal{M}\circ\Phi:\mathbb{R}^K\to\mathbb{R}^{|\mathcal{C}_K|}$ is definable, where $\mathcal{M}:\mathcal{A}_{|\mathcal{C}_K|}\to\mathbb{R}^{|\mathcal{C}_K|}$.
\end{lemma}
\begin{proof}

We establish a function in a composition way and show it is \textit{definable} in any $o$-minimal structures. 
According to Definition~\ref{defi_3}, an set $A\subset\mathbb{R}^n$ for some $n\in\mathbb{N}$ is called a \textit{definable set} if it satisfies the o-minimal structure. Naturally, for a definable set $A$, a map $a:A\to\mathbb{R}^r$ is said to be \textit{definable} if its graph is a definable set in $\mathbb{R}^{n+r}$. The simplest example of o-minimal structures is the family of semi-algebraic subsets of $\mathbb{R}^n$, which implies that the sublevel filtration is \textit{semi-algebraic}.
Since the PD can be regarded analytically as a function of filtration, we equivalently denote by $\mathcal{A}_{|\mathcal{C}_K|}$ the set of vectors in $\mathbb{R}^{|\mathcal{C}_K|}$, which defines a filtration in $\mathcal{C}_K$. We then use a persistence map $\mathcal{M}:\mathcal{A}_{|\mathcal{C}_K|}\to\mathbb{R}^{|\mathcal{C}_K|}$ to indicate the PD assigned to each filtration of $\mathcal{C}_K$, which consists of a permutation of the coordinates of $\mathbb{R}^{|\mathcal{C}_K|}$. Intuitively speaking, the permutation is a constant set of filtration such that it defines the same preorder on the cubes of $\mathcal{C}_K$. This leads to that the persistence map $\mathcal{M}$ is \textit{semi-algebraic}~\citep{alon2005crossing,zhao2024hausdorff} based on Proposition 3.2 in~\citep{carriere2021optimizing} and that there exists a semi-algebraic partition of $\mathcal{A}_{|\mathcal{C}_K|}$ such that the restriction of $\mathcal{M}$ to each element of each partition is a \textit{Lipschitz} map. Consequently, $\mathcal{M}\circ\Phi:\mathbb{R}^K\to\mathbb{R}^{|\mathcal{C}_K|}$ is definable.
\end{proof}

\noindent\textbf{Derivation of $\mathcal{L}_c$.} If $\Phi$ is differentiable, the differential of $\mathcal{M}\circ\Phi$ can trivially be computed through the partial derivatives of $\Phi$. As mentioned in the above, the connectivity loss is correlated to a function of persistence for a specific PD. Therefore, it can be defined as a function invariant to permutations of the points of the PD. We denote by $F$ the function of persistence such that if $F$ is locally Lipschitz, $F\circ\mathcal{M}$ is also \textit{locally Lipschitz}. Moreover, if $F$ is definable in an o-minimal structure, then for any definable filtration $\Phi$, the composition $\mathcal{L}_c=F\circ\mathcal{M}\circ\Phi: \mathbb{R}^K\to\mathbb{R}$ is also definable. 

$\mathcal{L}_c$ can be in a form of total distance of birth and death pairs of a PD. For example, for a PD denoted by $\mathcal{D}=((b_1,d_1),(b_2,d_2),...,(b_q,d_q))$, $\mathcal{L}_c(\mathcal{D})=\sum_{i=1}^q|b_i-d_i|$, which is obviously Lipschitz and definable in any o-minimal structure. 

\subsection{Additional proofs}
\textbf{\theoremref{theorem_1}}: Suppose that \assmref{assump_1} holds and that $\mathcal{L}_{sdf}$ is continuously differentiable.
    Let $\mathcal{C}_K$ be a cubical complex and $\Phi:\mathbb{R}^K\to\mathbb{R}^{|\mathcal{C}_K|}$ a family of filtrations of $\mathcal{C}_K$ that is definable in an $o$-minimal structure. Let $F:\mathbb{R}^{\mathcal{C}_K}\to\mathbb{R}$ be a definable function of persistence such that $\mathcal{L}_c=F\circ\mathcal{M}\circ\Phi$ is locally Lipschitz. Thus, the limit points of the sequence $\{\theta_t\}$ generated by \eqnref{subgradient_update} are critical points of $\mathcal{L}$ and the sequence $\mathcal{L}(\theta_t)$ converges to $\mathcal{L}^*$ almost surely.
\begin{proof}
    As $\mathcal{L}_c$ is continuously differentiable and $\mathcal{L}_{g}$ is continuously differentiable. It is immediately obtained that $\mathcal{L}=\mathcal{L}_{g}+\mathcal{L}_c$ continuously differentiable. As $\mathcal{L}$ is definable, it is locally Lipschitz. Following the proof closely from Corollary 5.9 in~\citep{davis2020stochastic}. The desirable conclusion is obtained.
\end{proof}

\begin{definition}($m\sim\epsilon$-dense set)\label{defi_5}
    Let $D\subset\mathbb{R}^3$ and $\epsilon>0$. $D$ is $\epsilon$-dense if and only if 
    \[\forall z\in D \exists z'\in D \setminus \{z\}:\|z-z'\|\leq \epsilon.\] For $m\in\mathbb{N}$, $D$ is $m\sim\epsilon$-dense if and only if:
    \[\exists M\subset D \setminus \{z\}:|M|=m,z'\in M\Rightarrow \|z-z'\|\leq \epsilon.\]
\end{definition}
Combining Definitions~\ref{defi_4} and~\ref{defi_5}, we then have the following result to provide insights into the density behavior of samples around points $z\in M$.
To mathematically characterize this, we introduce another definition to connect with the metric entropy~\citep{haussler1997mutual}.
\begin{definition}($\epsilon$-separated)\label{metric_entropy}
Let $D\subset\mathbb{R}^3,\epsilon>0$. $D$ is $\epsilon$-separated if and only if $\forall z, z'\in D: z\neq z'\Rightarrow \|z-z'\|\leq \epsilon$. For all $X\subset \mathbb{R}^3$. The $\epsilon$-metric entropy of $X$ is defined as
\begin{equation}
    \mathcal{N}_\epsilon(X)=\textnormal{max}\{|D|:D\subset X \textnormal\;{and}\;D\;\textnormal{is}\;\epsilon\textnormal{-separated}\}.
\end{equation}
\end{definition}
We further denote by $\mathcal{B}(z,\kappa)=\{z'\in\mathbb{R}^3:|z-z'|\leq \kappa\}$ the closed ball of radius $\kappa$ around $z$ such that its interior is immediately defined as $\mathcal{B}(z,\kappa)^0=\{z'\in\mathbb{R}^3:|z-z'|< \kappa\}$. Let $\mathcal{B}(z,\kappa,s)=\mathcal{B}(z,s) \backslash \mathcal{B}(z,\kappa)^0$ with $\kappa<s$ be the \textit{annulus} around $z$. Thus, the metric entropy of the annulus in $\mathbb{R}^3$ is defined as $\mathcal{E}^{\epsilon,3}_{\alpha,\beta}=\mathcal{N}_\epsilon(\mathcal{B}(0,\alpha,\beta))$. 

We next proceed to prove \theoremref{theorem_2} and \theoremref{theorem_3}. Before that, a technical lemma is presented.
\begin{lemma}\label{lemma_2}
    Let $2\leq k \leq m$ and $M\subset\mathbb{R}^3$ with $|M|=m$ such that for each $D\subset M$ with $|D|=k$, it holds that $D$ is $\alpha\sim\beta$-connected. Then for $v=m-k$ and any arbitrary but fixed $z\in M$, we find $M_z\subset M$ with $|M_z|=v+1$ and $M_z\subset B(z,\alpha,\beta)$.
\end{lemma}
\begin{proof}
    The proof follows similarly from Lemma 1 in~\citep{hofer2019connectivity}.
\end{proof}
With \lemmaref{lemma_2} in hand, we are now ready to show the proof for the following main results.

\noindent\textbf{\theoremref{theorem_2}}
    Let $2\leq k \leq m$ and $M\subset\mathbb{R}^3$ with $|M|=m$ such that for each $D\subset M$ with $|D|=k$, it holds that $D$ is $\alpha\sim\beta$-connected. Then $M$ is $(m-k+1)\sim\beta$-dense.
\begin{proof}
    By \lemmaref{lemma_2}, we can establish $m-k+1$ points, $M_z$, such that $M_z\subset \mathcal{B}(z,\alpha,\beta)$. Hence, we have the following relationship to hold true:
    \begin{equation}
        y\in M_z\Rightarrow \|z-y\|\leq \beta,
    \end{equation}
    which yields the desirable conclusion.
\end{proof}
\begin{remark}
Suppose that we have obtained the optimal $(b_i^*,d_i^*), \forall i\in[|\mathcal{C}_K|]$. When $\mathcal{L}$ converges, $\xi_i$ can be calculated in the last training epoch such that we can attain the approximate $\alpha$ and $\beta$. \theoremref{theorem_2} analytically quantifies the number of neighbors that can be found around $z\in M$ within distance $\beta$, which is $m-k+1$. Applying $M$ to $\mathbf{P}$ grants us the conclusion that $\mathbf{P}$ is $L-k+1\sim\beta$-dense. Note also that in this context, the conclusion applies to the whole dataset of $\mathbf{P}$, while in some scenarios, it can become only a batch of the dataset. Taking another different perspective of connectivity, we can study the separation of points in $M$. 
\end{remark}

\begin{theo}\label{theorem_3}
    Let $2\leq k \leq m$ and $M\subset\mathbb{R}^3$ with $|M|=m$ such that for each $D\subset M$ with $|D|=k$ such that $D$ is $\alpha\sim\beta$-connected. Then for $\epsilon>0$ and $m-k+1>\mathcal{E}^{\epsilon,3}_{\alpha,\beta}$, it follows that $M$ is not $\epsilon$-separated, where $\mathcal{E}^{\epsilon,3}_{\alpha,\beta}$ is defined in Definition~\ref{metric_entropy} in Appendix.
\end{theo}
\begin{proof}
    Let us choose some $z\in M$. By \lemmaref{lemma_2}, we can construct $m-k+1$ points, $M_z$, such that $M_z\subset \mathcal{B}(z,\alpha,\beta)$. The distance induced by the $l_2$ norm $\|\cdot\|$ is translation invariant. Thus, we have the following relationship:
    \begin{equation}
        \mathcal{E}^{\epsilon,3}_{\alpha,\beta}=\mathcal{N}_\epsilon(\mathcal{B}(z,\alpha,\beta)).
    \end{equation}
If $m-k+1>\mathcal{E}^{\epsilon,3}_{\alpha,\beta}$, it can be concluded that $M_z$ is not $\epsilon$-separated such that $M$ is not separated.
\end{proof}
According to \theoremref{theorem_3}, we can obtain a guideline to evaluate whether $\mathbf{P}$ is large enough to be no longer $\epsilon$-separated. It advocates that for each point in $\mathbf{P}$, there is no quantification for separatedness. Equivalently, we can only guarantee that with a size of $L$ for $\mathbf{P}$, there exist at least two points with distance smaller than $\epsilon$, which provides a necessary condition to ensure the connectivity on the topological surface for $\mathbf{P}$. 
In this study, we have used $\xi_i$ such that solving the optimization problem enables topological features to satisfy $b_i^*\neq d_i^*$ in $\varepsilon^*$. While in some scenarios, this can even be relaxed to have a threshold $\eta$ such that we will optimize $\xi_i=\eta-|b_i-d_i|$, which suggests that the optimization can be relatively easier to get features satisfying $b_i^*\neq d_i^*+
\eta$ or $d_i^*\neq b_i^*+
\eta$. Analogously, the conclusions from \theoremref{theorem_2} and \theoremref{theorem_3} can still apply herein by adjusting some constants.

\section{Additional Implementation Details}
\subsection{Architecture}
We utilize the same architecture as Neural-Pull's Pytorch implementation---8-layer MLP with 256 hidden dimensions, ReLU activation, and a geometric initialization scheme similar to IGR. We also use a skip connection in the middle layer.

\subsection{Training}
During training we randomly sample 20,000 points and randomly select 4096 query locations. We train for a total of 40,000 iterations, use the Adam optimizer with an initial learning rate of 0.001, and a cosine decay applied after the first 1000 iterations. Furthermore, we adopt a curriculum learning based approach for our proposed topological losses and introduce them in the last 500 iterations of the training process. We compute the cubical complex at a resolution of $16^3$. Note that Neural-Pull's PyTorch version utilizes an empirically determined scale parameter as opposed to a fixed value, utilized by the Tensorflow version. We run all our experiments on NVIDIA's A100 80GB GPU.

\subsection{Mesh extraction}
We utilize PyMCubes to reconstruct the mesh from the signed distance field at a resolution of $256^3$. 

\subsection{Evaluation metrics} We quantitatively evaluate our reconstructions using similar metrics as IGR~\citep{gropp2020implicit}. They are defined between two point sets, $P, Q \subset \mathbb{R}^3$---one-sided and two-sided Chamfer distance (CD) and Hausdorff distance (HD). We compute the distance of reconstructions to ground truth using the these distance metrics. They are defined as follows:
{\small
\begin{equation}
    {d}_{CD} (P,Q) = \frac{1}{2}(\vec{d_{CD}}(P,Q) + \vec{d_{CD}}(Q,P))
    \label{eq:chamfer_distance}
\end{equation}
\begin{equation}
    {d}_{HD} (P,Q) = \max\{\vec{d_{HD}}(P,Q), \vec{d_{HD}}(Q,P)\}
    \label{eq:hausdorff_distance}
\end{equation}}
where,
{\small
\begin{equation}
    \vec{d}_{CD} (P,Q) = \frac{1}{|P|}\sum_{{P_i}\in{P}}{\,\min_{{Q_j}\in{Q}}{{||{P_i}-{Q_j}}||}},
    \label{eq:one_sided_chamfer_distance}
\end{equation}
\begin{equation}
    \vec{d}_{HD} (P,Q) = \max_{{P_i}\in P}\min_{{Q_j}\in Q}{{||{P_i}-{Q_j}}||}
    \label{eq:one_sided_hausdorff_distance}
\end{equation}
}are the one-sided Chamfer and Hausdorff distances, respectively. 

Furthermore, to assess the effect of our proposed topological losses, we compute $\mathcal{L}_{\mathcal{S}}$ from the predicted signed distance field. This serves as a measure of preserving significant topological features, with lower magnitudes of the loss corresponding to high preservation of connected features.

\section{Ablation studies}
\label{sec:ablation}
We provide an ablation study of our proposed losses, geometric network initialization, originally proposed by \citet{atzmon2020sal}, and adopted by Neural-Pull.

\noindent\textbf{Topological loss contribution:} We begin by studying the effect of each of our proposed topological losses, $\mathcal{L}_{\mathcal{N}}$ and $\mathcal{L}_{\mathcal{S}}$ by testing on various shapes with thin structures. The loss contributions are shown in \tabref{tab:loss_ablation}. From the metrics, we see that a combined effect of $\mathcal{L}_{\mathcal{N}}$ and $\mathcal{L}_{\mathcal{S}}$ is necessary for us to capture relevant topological properties, as seen by the lowest mean two-sided CD of \textbf{0.0105} and reduced standard deviation of \textbf{0.0013}. In some cases (like the Eiffel Tower), a single loss function may suffice. This indicates that while the combined effect of both losses is beneficial overall, the optimal loss configuration may vary between geometries, given the unique characteristics of each shape.

\begin{table}[h!]
\centering
\caption{Comparison of two-sided CD (reconstruction and ground truth) for various thin structures using $\mathcal{L}_{\mathcal{S}}$ and $\mathcal{L}_{\mathcal{N}}$.}
\label{tab:loss_ablation}
\scriptsize
\setlength{\tabcolsep}{3pt}
\begin{tabular}{@{}l@{\hspace{4pt}}cccc@{}} 
\toprule
Model &  STITCH with $\mathcal{L}_{\mathcal{S}}$  & STITCH with $\mathcal{L}_{\mathcal{N}}$ & STITCH with  $\mathcal{L}_{\mathcal{N}}$ and $\mathcal{L}_{\mathcal{S}}$\\
\midrule
Torus        & 0.0162 & 0.0103 & \textbf{0.0084}\\
Hollow Ball  & 0.0171 & 0.0135 & \textbf{0.0107}\\
Eiffel Tower & \textbf{0.0087} & 0.0126 & 0.0121\\
Lamp         & 0.0196 & 0.0110 & \textbf{0.0107}\\
\midrule
Mean    & 0.0154 & 0.0119 & \textbf{0.0105}\\
Std.\,Dev. & 0.0041 & \textbf{0.0013} & \textbf{0.0013}\\
\bottomrule
\end{tabular}
\end{table}

\noindent\textbf{Geometric network initialization:}
We begin by evaluating the effect of the geometric network initialization (GNI) scheme and report our results on the thin structures dataset with and without GNI. STITCH achieves (see \tabref{tab:ablation_thin_gni}) a two-sided CD of \textbf{0.0105} and \textbf{0.0206}, with and without GNI, respectively, and we conclude that GNI helps our proposed framework learn better.

\begin{table}[h!]
\centering
\caption{Comparison of two-sided CD (reconstruction and input point cloud) for various thin structures.}
\label{tab:ablation_thin_gni}
\scriptsize
\setlength{\tabcolsep}{3pt}  
\begin{tabular}{@{}l@{\hspace{4pt}}cccccccc@{}} 
\toprule
Model & STITCH (no GNI) & STITCH (with GNI) \\
\midrule
Torus & 0.0202 & \textbf{0.0084}\\
Hollow Ball & 0.0146 & \textbf{0.0107}\\
Eiffel Tower & 0.0225 & \textbf{0.0121}\\
Lamp & 0.0252 & \textbf{0.0107}\\
\midrule
Mean    & 0.0206 & \textbf{0.0105}\\
Std.\,Dev. & 0.0039 & \textbf{0.0013}\\
\bottomrule
\end{tabular}
\end{table}

\begin{figure*}[!t]
    \centering
    \includegraphics[width=0.99\linewidth,trim={0.0in 1.0in 0.0in 1.0in},clip]{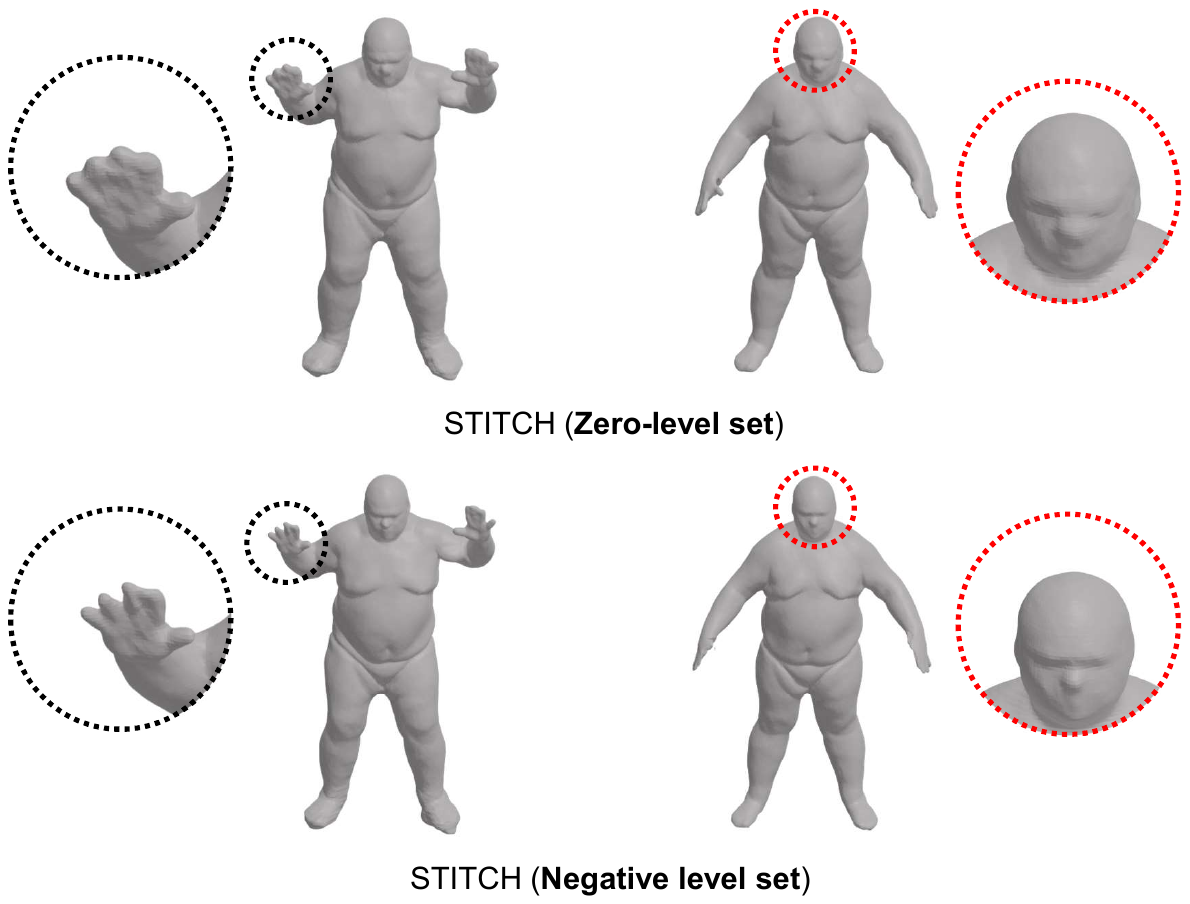}
    \caption{Images showing the details obtained by using a negative level set for DFAUST reconstructions.}
    \label{fig:dfaust_ablation}
\end{figure*}

\noindent\textbf{Effect of different level set:} 
Empirically, our approach tends to generate slightly inflated reconstructions for certain geometries. We primarily attribute this to the bias of our topological losses, which encourage a connected component on the zero-level set. In this ablation experiment, we consider a negative level set value of $\nicefrac{-1}{256}$ and test on the DFAUST models to see if that improves the accuracy metrics and the visual fidelity of the reconstructions. We find that the Chamfer distance is significantly lower for this level set value (see \tabref{tab:ablation_dfaust_two_sided_cd_pred_input}). We show the effect of the negative level set in \figref{fig:dfaust_ablation}.

\begin{table}[t!]
\centering
\caption{Comparison of two-sided CD (reconstruction and input point cloud) for five shapes from the DFAUST dataset.}
\label{tab:ablation_dfaust_two_sided_cd_pred_input}
\scriptsize
\setlength{\tabcolsep}{3pt}
\begin{tabular}{@{}l@{\hspace{4pt}}cccccccc@{}} 
\toprule
Model & STITCH (zero-level set) & STITCH (negative level set) \\
\midrule
Shape 1 & 0.0072 & \textbf{0.0047}\\
Shape 2 & 0.0070  & \textbf{0.0052}\\
Shape 3 & 0.0072 & \textbf{0.0049}\\
Shape 4 & 0.0070 & \textbf{0.0044}\\
Shape 5 & 0.0072 & \textbf{0.0050}\\
\midrule
Mean    & 0.0071 & \textbf{0.0048}\\
Std.\,Dev. & \textbf{0.0001} & 0.0003\\
\bottomrule
\end{tabular}
\end{table}

\begin{figure*}[!t]
    \centering
    \includegraphics[width=0.99\linewidth,trim={0.0in 1.0in 0.0in 1.0in},clip]{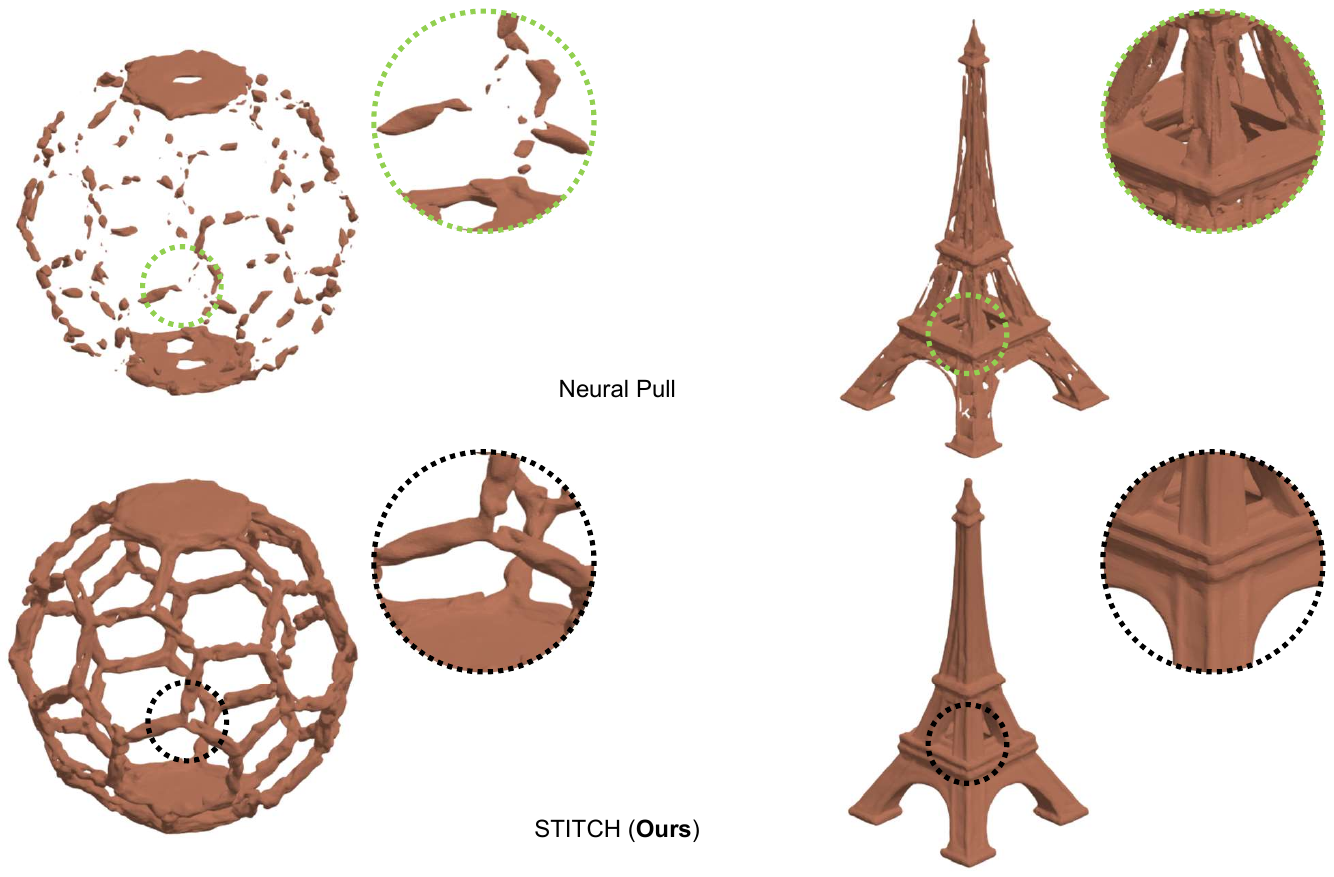}
    \caption{Images showing the details obtained by STITCH on thin objects.}
    \label{fig:thin_closeup}
\end{figure*}

\section{Additional Results}

We provide additional metrics on SRB~\cite{williams2019deep}, DFAUST~\cite{bogo2017dynamic}, Plants, and thin shapes datasets that were presented in our main paper. We also tested our approach on ABC and Famous datasets, but due to the new CVPR policy on presenting additional dataset results in the supplement, we do not include them.

SRB dataset provides a ``scan'' point cloud (normals included) that simulates noisy range scanning with missing regions and misalignment. They also provide a ``ground truth'' point cloud (densely covers the surface without any missing regions). For our experiments, by ``input point cloud'' we refer to the ``scan'' and ``ground truth points'' refer to the ``ground truth''. We showcase the significant feature topological loss (\tabref{tab:srb_tda_metrics}), one-sided CD (\tabref{tab:srb_one_sided_pred_gt}, \tabref{tab:srb_one_sided_cd_gt_pred}, \tabref{tab:srb_one_sided_cd_input_pred}, and \tabref{tab:srb_one_sided_cd_pred_input}), two-sided CD (\tabref{tab:srb_two_sided_cd_pred_input}, \tabref{tab:srb_two_sided_cd_pred_gt}), and two-sided HD (\tabref{tab:srb_two_sided_hd_pred_gt} and \tabref{tab:srb_two_sided_hd_pred_input}). STITCH exhibits consistent performance, evident by the low standard deviations across metrics, thus making it reliable across diverse shapes in SRB. For instance, STITCH achieves a mean two-sided CD of \textbf{0.0084} with a standard deviation of \textbf{0.0014}, indicating at-par performance with other neural baselines, and shows balanced performance when evaluated against input scans and ground truth surfaces.

The DFAUST dataset provides raw scans (point clouds) and their ground truth mesh counterparts. In our experiments, we refer to the ``raw scan'' as ``input point cloud.'' We report the significant feature topological loss (\tabref{tab:dfaust_tda_metrics}), one-sided CD (\tabref{tab:dfaust_one_sided_cd_input_pred}, \tabref{tab:dfaust_one_sided_cd_pred_input}), two-sided CD (\tabref{tab:dfaust_two_sided_cd_pred_input}) and two-sided HD (\tabref{tab:dfaust_two_sided_hd_pred_input}). STITCH excels in preserving significant topological features by achieving the lowest significant feature loss among all methods for almost all shapes. Furthermore, STITCH maintains the lowest variability in both CD and HD, with a standard deviation as low as \textbf{0.0003} in one-sided CD. We note that our reconstructions for DFAUST are slightly inflated compared to the expected shape and attribute this to the bias of our proposed topological losses to encourage a connected component on the zero-level set. To this end, as presented above, we performed an ablation study that looks at the non-zero level set to get better-looking reconstructions.

As for the thin objects dataset, we also report the significant feature topological loss (\tabref{tab:thin_tda_metrics}), one-sided CD (\tabref{tab:thin_one_sided_cd_pred_gt}, \tabref{tab:thin_one_sided_cd_gt_pred}), two-sided CD (\tabref{tab:thin_two_sided_cd_pred_gt}), and two-sided HD (\tabref{tab:thin_two_sided_hd_pred_gt}). STITCH effectively preserves the topological properties of thin structures, achieving competitive significant feature loss values (e.g. \textbf{0.5299} for Eiffel Tower) and low CD and HD metrics, further validating its capability in handling complex geometries. We showcase closeups of thin geometries in \figref{fig:thin_closeup}.

\begin{figure*}[!t]
    \centering
    \includegraphics[width=0.99\linewidth,trim={0.0in 1.0in 0.0in 1.0in},clip]{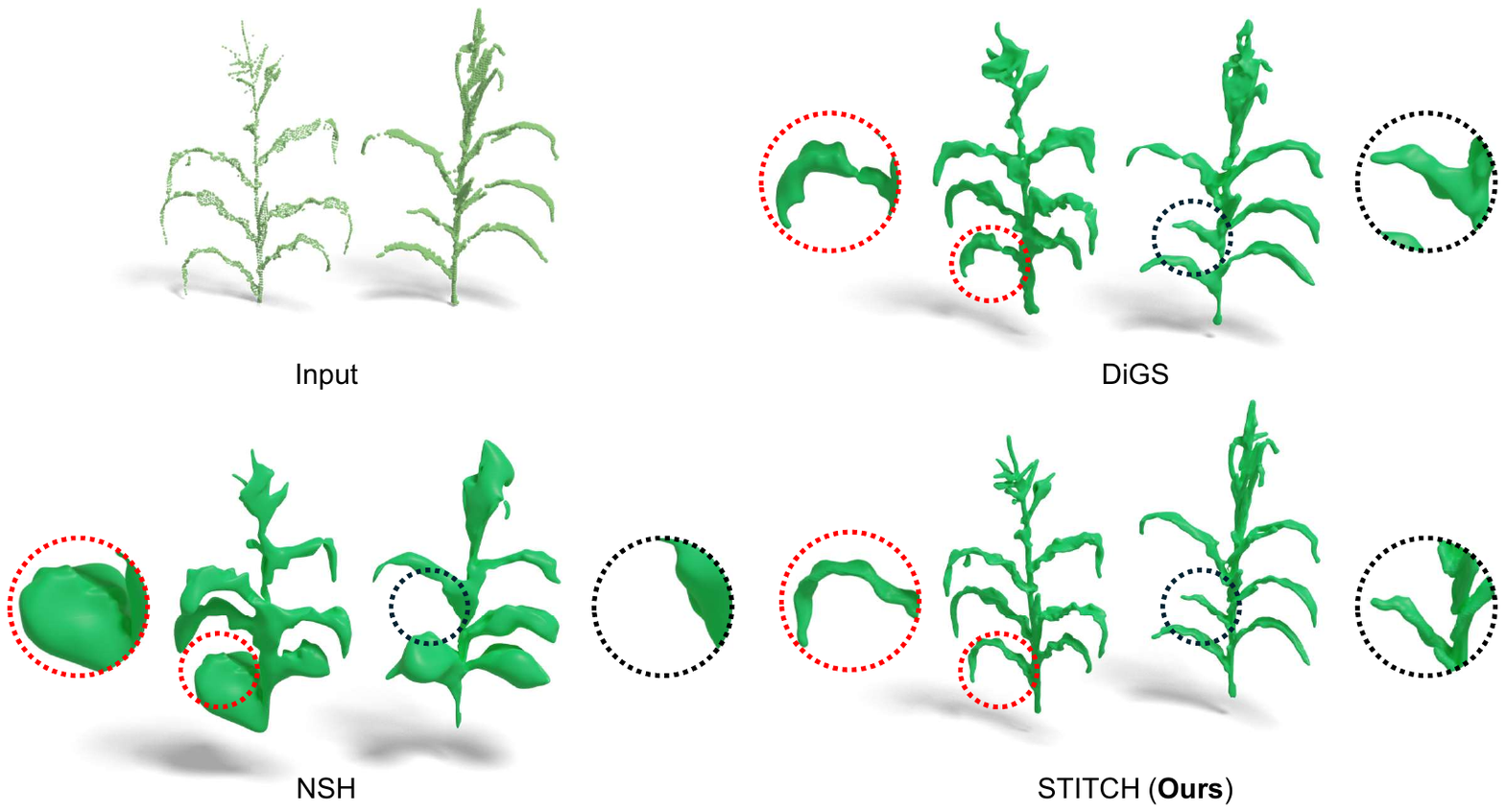}
    \caption{Images showing the details of two plant reconstructions.}
    \label{fig:plant_closeup}
\end{figure*}

Finally, on the Plants dataset, we report the significant feature topological loss (\tabref{tab:plant_tda_metrics}), one-sided CD (\tabref{tab:plants_one_sided_cd_gt_pred}), two-sided CD (\tabref{tab:plants_two_sided_cd_pred_gt}), and two-sided HD (\tabref{tab:plants_two_sided_hd_pred_gt}). STITCH outperforms all other baselines in significant feature loss, demonstrating its ability to retain crucial geometric details relevant to slender plant structures. Furthermore, STITCH consistently ranks near the top in CD and HD metrics and exhibits low standard deviations, showing balanced performance. We showcase closeups of plant geometries in \figref{fig:plant_closeup}.

Overall, STITCH demonstrates strong and consistent performance across various datasets, particularly excelling in preserving significant topological features while maintaining low variability in geometric error metrics.


\begin{table}[t!]
\centering
\caption{Comparison of one-sided CD (ground truth points to reconstruction) for models from the SRB.}
\label{tab:srb_one_sided_cd_gt_pred}
\scriptsize
\setlength{\tabcolsep}{3pt}
\begin{tabular}{@{}l@{\hspace{4pt}}cccccccc@{}} 
\toprule
Model & PSR & NP & IGR & DiGS & OG-INR & NSH & STITCH (Ours) \\
\midrule
Anchor    & 0.0104 & 0.0099 &  -     & 0.0113 & \textbf{0.009}  & 0.0094 & 0.0100\\
Daratech  & 0.0066 & 0.0123 &  -     & 0.0073 & 0.0064 & \textbf{0.0061} & 0.0100\\
Dc        & 0.0065 & 0.0141 & 0.227  & 0.0080  & \textbf{0.0062} & \textbf{0.0062} & 0.0080\\
Gargoyle  & 0.0083 & 0.0103 &  -     & 0.0095 & 0.0083 & \textbf{0.0075} & 0.0100\\
Lord\,Quas & 0.0047 & 0.0063 & 0.1444 & 0.0063 & 0.0047 & \textbf{0.0046} & 0.0060\\
\midrule
Mean       & 0.0073 & 0.0106 & -      & 0.0085 & 0.0069 & \textbf{0.0068} & 0.0088 \\
Std.\,Dev.    & 0.0019 & 0.0026 & -      & 0.0018 & \textbf{0.0015} & 0.0016 & 0.0016 \\
\bottomrule
\end{tabular}
\end{table}

\begin{table}[t!]
\centering
\caption{Comparison of one-sided CD (input point cloud to reconstruction) for models from the SRB.}
\label{tab:srb_one_sided_cd_input_pred}
\scriptsize
\setlength{\tabcolsep}{3pt}
\begin{tabular}{@{}l@{\hspace{4pt}}cccccccc@{}} 
\toprule
Model & PSR & NP & IGR & DiGS & OG-INR & NSH & STITCH (Ours) \\
\midrule
Anchor    & \textbf{0.0063} & 0.0089 &   -    & 0.0090 & 0.0066 & 0.0065 & 0.0091\\
Daratech  & \textbf{0.0046} & 0.0100 &   -    & 0.0064 & 0.0049 & 0.0053 & 0.0088\\
Dc        & \textbf{0.0049} & 0.0142 & 0.2488 & 0.0072 & 0.0051 & 0.0052 & 0.0075\\
Gargoyle  & \textbf{0.0056} & 0.0082 &   -    & 0.0085 & 0.0059 & 0.0061 & 0.0082\\
Lord\,Quas & \textbf{0.0038} & 0.0057 & 0.1484 & 0.0057 & 0.0039 & 0.0041 & 0.0057\\
\midrule
Mean       & \textbf{0.0050} & 0.0094 & -      & 0.0074 & 0.0053 & 0.0054 & 0.0079 \\
Std.\,Dev.    & \textbf{0.0009} & 0.0028 & -      & 0.0012 & 0.0009 & 0.0008 & 0.0012 \\
\bottomrule
\end{tabular}
\end{table}

\begin{table}[t!]
\centering
\caption{Comparison of one sided CD (reconstruction to input point cloud) for models from the SRB.}
\label{tab:srb_one_sided_cd_pred_input}
\scriptsize
\setlength{\tabcolsep}{3pt}
\begin{tabular}{@{}l@{\hspace{4pt}}cccccccc@{}} 
\toprule
Model & PSR & NP & IGR & DiGS & OG-INR & NSH & STITCH (Ours) \\
\midrule
Anchor    & \textbf{0.0063} & 0.0118 &   -    & 0.011  & 0.0096 & 0.0080 & 0.0116\\
Daratech  & \textbf{0.0047} & 0.007  &   -    & 0.0079 & 0.0056 & 0.0060 & 0.0090\\
Dc        & \textbf{0.0052} & 0.0089 & 0.2269 & 0.0088 & 0.0065 & 0.0064 & 0.0087\\
Gargoyle  & \textbf{0.0059} & 0.0090 &   -    & 0.0093 & 0.0068 & 0.0067 & 0.0091\\
Lord\,Quas & \textbf{0.004}  & 0.0062 & 0.0856 & 0.0066 & 0.0046 & 0.0046 & 0.0062\\
\midrule
Mean       & \textbf{0.0052} & 0.0086 & -      & 0.0087 & 0.0066 & 0.0063 & 0.0089 \\
Std.\,Dev.    & \textbf{0.0008} & 0.0019 & -      & 0.0015 & 0.0017 & 0.0011 & 0.0017 \\
\bottomrule
\end{tabular}
\end{table}

\begin{table}[t!]
\centering
\caption{Comparison of two sided CD (reconstruction and input point cloud) for models from the SRB.}
\label{tab:srb_two_sided_cd_pred_input}
\scriptsize
\setlength{\tabcolsep}{3pt}
\begin{tabular}{@{}l@{\hspace{4pt}}cccccccc@{}} 
\toprule
Model & PSR & NP & IGR & DiGS & OG-INR & NSH & STITCH (Ours) \\
\midrule
Anchor    & \textbf{0.0063} & 0.0104 & -  & 0.0100 & 0.0124 & 0.0096 & 0.0103\\
Daratech  & \textbf{0.0047} & 0.0085 & - & 0.0072 & 0.0090 & 0.0072 & 0.0089\\
Dc        & \textbf{0.0051} & 0.0115 & 0.2379  & 0.0080 & 0.0110 & 0.0079 & 0.0081\\
Gargoyle  & \textbf{0.0057} & 0.0086 & - & 0.0089 & 0.0117 & 0.0087 & 0.0086\\
Lord\,Quas & \textbf{0.0039} & 0.0059 & 0.1170 & 0.0061 & 0.0086 & 0.0061 & 0.0059\\
\midrule
Mean       & \textbf{0.0051} & 0.0090 & - & 0.0080 & 0.0105 & 0.0079 & 0.0084 \\
Std.\,Dev.    & \textbf{0.0008} & 0.0019 & - & 0.0013 & 0.0015 & 0.0012 & 0.0014 \\
\bottomrule
\end{tabular}
\end{table}

\begin{table}[t!]
\centering
\caption{Comparison of two sided CD (reconstruction and ground truth points) for models from the SRB.}
\label{tab:srb_two_sided_cd_pred_gt}
\scriptsize
\setlength{\tabcolsep}{3pt}
\begin{tabular}{@{}l@{\hspace{4pt}}cccccccc@{}} 
\toprule
Model & PSR & NP & IGR & DiGS & OG-INR & NSH & STITCH (Ours) \\
\midrule
Anchor    & \textbf{0.0090} & 0.0098 & - & 0.0108 & 0.0134 & 0.0108 & 0.0097\\
Daratech  & \textbf{0.0062} & 0.0097 & - & 0.0073 & 0.0104 & 0.0074 & 0.0095\\
Dc        & \textbf{0.0063} & 0.0110 & 0.2282 & 0.0080 & 0.0117 & 0.0080 & 0.0082\\
Gargoyle  & \textbf{0.0080} & 0.0101 & - & 0.0095 & 0.0104 & 0.0094 & 0.0103\\
Lord\,Quas & \textbf{0.0047} & 0.0063 & 0.1148 & 0.0063 & 0.0079 & 0.0062 & 0.0063\\
\midrule
Mean       & \textbf{0.0068} & 0.0094 & - & 0.0084 & 0.0108 & 0.0084 & 0.0088\\
Std.\,Dev. & 0.0015 & 0.0016 & - & 0.0016 & 0.0018 & 0.0016 & \textbf{0.0014}\\
\bottomrule
\end{tabular}
\end{table}

\begin{table}[t!]
\centering
\caption{Comparison of two sided HD (reconstruction and ground truth points) for models from the SRB.}
\label{tab:srb_two_sided_hd_pred_gt}
\scriptsize
\setlength{\tabcolsep}{3pt}
\begin{tabular}{@{}l@{\hspace{4pt}}cccccccc@{}} 
\toprule
Model & PSR & NP & IGR & DiGS & OG-INR & NSH & STITCH (Ours) \\
\midrule
Anchor    & 0.1066 & 0.0699 & - & 0.0985 & 0.0963 & 0.1006 & \textbf{0.0654}\\
Daratech  & 0.0845 & 0.0667 & - & 0.0429 & \textbf{0.0357} & 0.0399 & 0.0841\\
Dc        & 0.0415 & 0.1035 & 0.654 & \textbf{0.0297} & 0.0331 & 0.0312 & 0.0458\\
Gargoyle  & 0.0622 & 0.0642 & - & \textbf{0.0548} & 0.0642 & 0.0619 & 0.0581\\
Lord\,Quas & 0.0259 & 0.0337 & 0.4206 & 0.0268 & 0.0261 & \textbf{0.0198} & 0.0351\\
\midrule
Mean       & 0.0641 & 0.0676 & - & \textbf{0.0505} & 0.0511 & 0.0507 & 0.0577\\
Std.\,Dev. & 0.0290 & 0.0222 & - & 0.0260 & 0.0261 & 0.0285 & \textbf{0.0168}\\
\bottomrule
\end{tabular}
\end{table}

\begin{table}[t!]
\centering
\caption{Comparison of two sided HD (reconstruction and input point cloud) for models from the SRB.}
\label{tab:srb_two_sided_hd_pred_input}
\scriptsize
\setlength{\tabcolsep}{3pt}
\begin{tabular}{@{}l@{\hspace{4pt}}cccccccc@{}} 
\toprule
Model & PSR & NP & IGR & DiGS & OG-INR & NSH & STITCH (Ours) \\
\midrule
Anchor    & \textbf{0.0721} & 0.1189 & - & 0.1003 & 0.1161 & 0.1039 & 0.1132\\
Daratech  & 0.1011 & 0.0614 & - & 0.0658 & \textbf{0.0347} & 0.0423 & 0.0819\\
Dc        & \textbf{0.0502} & 0.1018 & 0.6493 & 0.0556 & 0.0593 & 0.0547 & 0.0527\\
Gargoyle  & \textbf{0.0433} & 0.0596 & - & 0.0636 & 0.0647 & 0.0528 & 0.0637\\
Lord\,Quas & \textbf{0.0452} & 0.0548 & 0.4191 & 0.0614 & 0.0563 & 0.0581 & 0.0492\\
\midrule
Mean       & \textbf{0.0624} & 0.0793 & - & 0.0693 & 0.0662 & \textbf{0.0624} & 0.0718\\
Std.\,Dev. & 0.0219 & 0.0260 & - & \textbf{0.0158} & 0.0269 & 0.0214 & 0.0228\\
\bottomrule
\end{tabular}
\end{table}


\begin{table}[t!]
\centering
\caption{Comparison of significant features topological loss term for models from DFAUST dataset.}
\label{tab:dfaust_tda_metrics}
\scriptsize
\setlength{\tabcolsep}{3pt}
\begin{tabular}{@{}l@{\hspace{4pt}}ccccc@{}} 
\toprule
Model & NP & DiGS & OG-INR & STITCH (Ours) \\
\midrule
Shape 1 & \textbf{1.9789} & 8.1401 & 112.1125 & 4.2243\\
Shape 2 & 3.4147 & 9.7166 & 26.8527 & \textbf{0.1238}\\
Shape 3 & \textbf{4.0755} & 6.5721 & 19.5115 & 4.7476\\
Shape 4 & 18.5675 & 6.0786 & 11.9304 & \textbf{0.2747}\\
Shape 5 & 3.1574 & 7.7776 & 11.9195 & \textbf{0.9108}\\
\bottomrule
\end{tabular}
\end{table}

\begin{table}[t!]
\centering
\caption{Comparison of one-sided CD (input point cloud to reconstruction) for five shapes from the DFAUST dataset.}
\label{tab:dfaust_one_sided_cd_input_pred}
\scriptsize
\setlength{\tabcolsep}{3pt}
\begin{tabular}{@{}l@{\hspace{4pt}}cccccccc@{}} 
\toprule
Model & PSR & NP & IGR & DiGS & OG-INR & NSH & STITCH (Ours) \\
\midrule
Shape 1 & 0.1486 & 0.0078 & 0.0178 & 0.0039 & \textbf{0.0033} & \textbf{0.0033} & 0.0062\\
Shape 2 & 0.1770 & 0.0168 & 0.0265 & \textbf{0.0034} & \textbf{0.0034} & \textbf{0.0034} & 0.0065\\
Shape 3 & 0.1722 & 0.0112 & 0.0077 & \textbf{0.0041} & 0.0043 & 0.0053 & 0.0061\\
Shape 4 & 0.0870 &   -    & 0.0052 & 0.0055 & \textbf{0.0031} & 0.0033 & 0.0067\\
Shape 5 & 0.1463 & 0.0043 & 0.0071 & 0.0037 & \textbf{0.0034} & 0.0037 & 0.0068\\
\midrule
Mean    & 0.1462 &   -    & 0.0129 & 0.0041 & \textbf{0.0035} & 0.0038 & 0.0065 \\
Std.\,Dev. & 0.0320 &   -    & 0.0081 & 0.0007 & 0.0004 & 0.0008 & \textbf{0.0003} \\
\bottomrule
\end{tabular}
\end{table}

\begin{table}[t!]
\centering
\caption{Comparison of one-sided CD (reconstruction to input point cloud) for five shapes from the DFAUST dataset.}
\label{tab:dfaust_one_sided_cd_pred_input}
\scriptsize
\setlength{\tabcolsep}{3pt}
\begin{tabular}{@{}l@{\hspace{4pt}}cccccccc@{}} 
\toprule
Model & PSR & NP & IGR & DiGS & OG-INR & NSH & STITCH (Ours) \\
\midrule
Shape 1 & 0.2611 & 0.0077 & 0.0178 & 0.0042 & 0.0037 & \textbf{0.0036} & 0.0082\\
Shape 2 & 0.2911 & 0.0177 & 0.0275 & 0.0040 & 0.0040 & \textbf{0.0038} & 0.0074\\
Shape 3 & 0.2538 & \textbf{0.0041} & 0.0091 & 0.0044 & 0.0046 & 0.0055 & 0.0084\\
Shape 4 & 0.1200 &   -    & 0.0113 & 0.0058 & \textbf{0.0035} & 0.0037 & 0.0073\\
Shape 5 & 0.2915 & 0.0047 & 0.0074 & 0.0039 & \textbf{0.0038} & 0.0039 & 0.0076\\
\midrule
Mean    & 0.2435 & -      & 0.0146 & 0.0045 & \textbf{0.0039} & 0.0041 & 0.0078 \\
Std.\,Dev. & 0.0636 & -      & 0.0073 & 0.0007 & \textbf{0.0004} & 0.0007 & \textbf{0.0004} \\
\bottomrule
\end{tabular}
\end{table}

\begin{table}[t!]
\centering
\caption{Comparison of two-sided CD (reconstruction and input point cloud) for five shapes from the DFAUST dataset.}
\label{tab:dfaust_two_sided_cd_pred_input}
\scriptsize
\setlength{\tabcolsep}{3pt}
\begin{tabular}{@{}l@{\hspace{4pt}}cccccccc@{}} 
\toprule
Model & PSR & NP & IGR & DiGS & OG-INR & NSH & STITCH (Ours) \\
\midrule
Shape 1 & 0.2048 & 0.0077 & 0.0178 & 0.0041 & 0.0035 & \textbf{0.0034} & 0.0072\\
Shape 2 & 0.2340 & 0.0173 & 0.0270 & 0.0037 & 0.0037 & \textbf{0.0036} & 0.0070\\
Shape 3 & 0.2130 & 0.0076 & 0.0084 & \textbf{0.0042} & 0.0045 & 0.0054 & 0.0072\\
Shape 4 & 0.1035 & - & 0.0083 & 0.0057 & \textbf{0.0033} & 0.0035 & 0.0070\\
Shape 5 & 0.2189 & 0.0045 & 0.0073 & 0.0038 & \textbf{0.0036} & 0.0038 & 0.0072\\
\midrule
Mean    & 0.1948 & - & 0.0138 & 0.0043 & \textbf{0.0037} & 0.0039 & 0.0071\\
Std.\,Dev. & 0.0467 & - & 0.0076 & 0.0007 & 0.0004 & 0.0007 & \textbf{0.0001}\\
\bottomrule
\end{tabular}
\end{table}

\begin{table}[t!]
\centering
\caption{Comparison of two-sided HD (reconstruction and input point cloud) for five shapes from the DFAUST dataset.}
\label{tab:dfaust_two_sided_hd_pred_input}
\scriptsize
\setlength{\tabcolsep}{3pt}  
\begin{tabular}{@{}l@{\hspace{4pt}}cccccccc@{}} 
\toprule
Model & PSR & NP & IGR & DiGS & OG-INR & NSH & STITCH (Ours) \\
\midrule
Shape 1 & 0.8346 & 0.0739 & 0.0626 & 0.0342 & 0.0340 & 0.0341 & \textbf{0.0239}\\
Shape 2 & 0.8680 & 0.0742 & 0.1377 & 0.0384 & \textbf{0.0336} & 0.0365 & 0.0353\\
Shape 3 & 0.8517 & 0.2641 & 0.0973 & 0.0310 & 0.0300 & 0.0298 & \textbf{0.0232}\\
Shape 4 & 0.6781 & - & 0.1626 & 0.0359 & 0.0329 & \textbf{0.0324} & 0.0437\\
Shape 5 & 0.8683 & 0.0361 & 0.0441 & \textbf{0.0326} & 0.0351 & 0.0348 & 0.0357\\
\midrule
Mean    & 0.8201 & - & 0.1009 & 0.0344 & 0.0331 & 0.0335 & \textbf{0.0324}\\
Std.\,Dev. & 0.0721 & - & 0.0444 & 0.0026 & \textbf{0.0017} & 0.0023 & 0.0078\\
\bottomrule
\end{tabular}
\end{table}


\begin{table}[t!]
\centering
\caption{Comparison of significant features topological loss term for various thin structures.}
\label{tab:thin_tda_metrics}
\scriptsize
\setlength{\tabcolsep}{3pt}
\begin{tabular}{@{}l@{\hspace{4pt}}ccccc@{}} 
\toprule
Model & NP & DiGS & OG-INR & STITCH (Ours) \\
\midrule
Torus & 21.2374 & 99.3480 & \textbf{9.8409} & 11.8474\\
Hollow Ball & 13.1609 & 132.5943 & 28.0065 & \textbf{6.2974}\\
Eiffel Tower & 3.0353 & 32.3211 & 12.8850 & \textbf{0.5299}\\
Lamp & 15.6391 & 38.3974 & 27.5075 & \textbf{11.1135}\\

\bottomrule
\end{tabular}
\end{table}

\begin{table}[t!]
\centering
\caption{Comparison of one-sided CD (reconstruction to ground truth point cloud) for various thin structures.}
\label{tab:thin_one_sided_cd_pred_gt}
\scriptsize
\setlength{\tabcolsep}{3pt}
\begin{tabular}{@{}l@{\hspace{4pt}}cccccccc@{}} 
\toprule
Model & PSR & NP & IGR & DiGS & OG-INR & NSH & STITCH (Ours) \\
\midrule
Torus  & - & 0.0065 & - & \textbf{0.0051} & 0.0097 & 0.0096 & 0.0088\\
Hollow Ball  & 0.0495 & \textbf{0.0109} & - & 0.0407 & 0.0359 & 0.0353 & 0.0124\\
Eiffel Tower  & 0.0081 & \textbf{0.0080} & - & 0.0084 & \textbf{0.0080} & 0.0081 & 0.0123\\
Lamp  & 0.0283 & \textbf{0.0098} & - & 0.0155 & 0.0196 & 0.0198 & 0.0121\\
\midrule
Mean    & 0.0286 & \textbf{0.0088} & - & 0.0174 & 0.0183 & 0.0182 & 0.0114\\
Std.\,Dev. & 0.0169 & 0.0017 & - & 0.0140 & 0.0111 & 0.0109 & \textbf{0.0015}\\
\bottomrule
\end{tabular}
\end{table}

\begin{table}[t!]
\centering
\caption{Comparison of one-sided CD (ground truth point cloud to reconstruction) for various thin structures.}
\label{tab:thin_one_sided_cd_gt_pred}
\scriptsize
\setlength{\tabcolsep}{3pt}
\begin{tabular}{@{}l@{\hspace{4pt}}cccccccc@{}} 
\toprule
Model & PSR & NP & IGR & DiGS & OG-INR & NSH & STITCH (Ours) \\
\midrule
Torus & - & 0.0176 & - & \textbf{0.0051} & 0.0077 & 0.0076 & 0.0079\\
Hollow Ball & 0.0293 & 0.0136 & - & 0.0113 & 0.0089 & \textbf{0.0088} & 0.0090\\
Eiffel Tower & 0.0064 & \textbf{0.0047} & - & 0.0051 & \textbf{0.0047} & \textbf{0.0047} & 0.0120\\
Lamp & 0.0191 & 0.0141 & - & 0.0091 & 0.0090 & \textbf{0.0089} & 0.0093\\
\midrule
Mean    & 0.0183 & 0.0125 & - & 0.0077 & 0.0076 & \textbf{0.0075} & 0.0096\\
Std.\,Dev. & 0.0094 & 0.0048 & - & 0.0027 & 0.0017 & 0.0017 & \textbf{0.0015}\\
\bottomrule
\end{tabular}
\end{table}

\begin{table}[t!]
\centering
\caption{Comparison of two-sided CD (reconstruction and ground truth point cloud) for various thin structures.}
\label{tab:thin_two_sided_cd_pred_gt}
\scriptsize
\setlength{\tabcolsep}{3pt}
\begin{tabular}{@{}l@{\hspace{4pt}}cccccccc@{}} 
\toprule
Model & PSR & NP & IGR & DiGS & OG-INR & NSH & STITCH (Ours) \\
\midrule
Torus        & -      & 0.0121 & -      & 0.0051 & 0.0087 & 0.0086 & \textbf{0.0084}\\
Hollow Ball  & 0.0394 & 0.0123 & - & 0.0260 & 0.0224 & 0.0221 & \textbf{0.0107}\\
Eiffel Tower & 0.0073 &   \textbf{0.0063}    &   -     &   0.0068     &  0.0064      &  0.0064      & 0.0121\\
Lamp         & 0.0237 &  0.0120     & -       &  0.0123      & 0.0143       &   0.0143     & \textbf{0.0107}\\
\midrule
Mean    & 0.0235 & 0.0107 & - & 0.0126 & 0.0130 & 0.0129 & \textbf{0.0105}\\
Std.\,Dev. & 0.0131 & \textbf{0.0025} & - & 0.0082 & 0.0062 & 0.0061 & 0.0013\\
\bottomrule
\end{tabular}
\end{table}

\begin{table}[t!]
\centering
\caption{Comparison of two-sided HD (reconstruction and ground truth point cloud) for various thin structures.}
\label{tab:thin_two_sided_hd_pred_gt}
\scriptsize
\setlength{\tabcolsep}{3pt}
\begin{tabular}{@{}l@{\hspace{4pt}}cccccccc@{}} 
\toprule
Model & PSR & NP & IGR & DiGS & OG-INR & NSH & STITCH (Ours) \\
\midrule
Torus  & - & 0.0810 & - & 0.2575 & 0.0376 & 0.0378 & \textbf{0.0266}\\
Hollow Ball  & 0.4503 & 0.0666 & - & 0.1603 & 0.1956 & 0.1966 & \textbf{0.0589}\\
Eiffel Tower  & 0.0432 & \textbf{0.0383} & - & 0.0392 & 0.0370 & 0.0384 & 0.0455\\
Lamp  & 0.0841 & 0.0707 & - & 0.1658 & 0.1303 & 0.1280 & \textbf{0.0470}\\
\midrule
Mean    & 0.1925 & 0.0642 & - & 0.1557 & 0.1001 & 0.1002 & \textbf{0.0445}\\
Std.\,Dev. & 0.1830 & 0.0158 & - & 0.0776 & 0.0669 & 0.0667 & \textbf{0.0116}\\
\bottomrule
\end{tabular}
\end{table}



\begin{table}[t!]
\centering
\caption{Comparison of significant features topological loss term for eight different plants.}
\label{tab:plant_tda_metrics}
\scriptsize
\setlength{\tabcolsep}{3pt}
\begin{tabular}{@{}l@{\hspace{4pt}}ccccc@{}}
\toprule
Model & NP & DiGS & OG-INR & STITCH (Ours) \\
\midrule
Plant 1 & 2.9234 & 4.7997 & 7.7186 & \textbf{0.9662}\\
Plant 2 & 1.7810 & 3.5856 & 12.0813 & \textbf{0.7643}\\
Plant 3 & 3.2288 & 3.2891 & 10.8686 & \textbf{0.5635}\\
Plant 4 & 3.9293 & 3.3183 & 9.1537 & \textbf{0.5263}\\
Plant 5 & 3.6036 & 2.0710 & 8.9453 & \textbf{0.9284}\\
Plant 6 & 5.4292 & 4.6589 & 10.4360 & \textbf{0.8159}\\
Plant 7 & 5.1283 & 3.4163 & 5.7550 & \textbf{1.0232}\\
Plant 8 & 3.8545 & 5.2666 & 5.5185 & \textbf{1.4380}\\
\bottomrule
\end{tabular}
\end{table}

\begin{table}[t!]
\centering
\caption{Comparison of one-sided CD (ground truth point cloud to reconstruction) for eight different plants.}
\label{tab:plants_one_sided_cd_gt_pred}
\scriptsize
\setlength{\tabcolsep}{3pt}
\begin{tabular}{@{}l@{\hspace{4pt}}cccccccc@{}} 
\toprule
Model & PSR & NP & IGR & DiGS & OG-INR & NSH & STITCH (Ours) \\
\midrule
Plant 1 & 0.0755 & - & 0.1210 & \textbf{0.0036} & - & 0.0052 & 0.0053\\
Plant 2 & 0.1677 & 0.0340 & - & 0.0059 & - & 0.0092 & \textbf{0.0047}\\
Plant 3 & 0.0971 & - & - & \textbf{0.0029} & 0.0037 & 0.0067 & 0.0046\\
Plant 4 & 0.1212 & - & - & 0.0038 & \textbf{0.0034} & 0.0057 & 0.0046\\
Plant 5 & 0.1362 & - & 0.2627 & 0.0045 & \textbf{0.0033} & 0.0226 & 0.0046\\
Plant 6 & 0.1104 & - & - & \textbf{0.0046} & 0.0047 & 0.0075 & 0.0051\\
Plant 7 & 0.0851 & - & - & 0.0058 & \textbf{0.0033} & 0.0155 & 0.0053\\
Plant 8 & 0.0717 & - & 0.2323 & \textbf{0.0040} & - & 0.0149 & 0.0046\\
\midrule
Mean    & 0.1081 & - & - & \textbf{0.0044} & - & 0.0109 & 0.0049\\
Std.\,Dev. & 0.0307 & - & - & 0.0010  & - & 0.0058 & \textbf{0.0003}\\
\bottomrule
\end{tabular}
\end{table}

\begin{table}[t!]
\centering
\caption{Comparison of two-sided CD (reconstruction and ground truth point cloud) for eight different plants.}
\label{tab:plants_two_sided_cd_pred_gt}
\scriptsize
\setlength{\tabcolsep}{3pt}
\begin{tabular}{@{}l@{\hspace{4pt}}cccccccc@{}} 
\toprule
Model & PSR & NP & IGR & DiGS & OG-INR & NSH & STITCH (Ours) \\
\midrule
Plant 1 & 0.1564 & - & 0.3186 & \textbf{0.0073} & - & 0.0138 & 0.0092\\
Plant 2 & 0.2077 & 0.0487 & - & 0.0095 & - & 0.0199 & \textbf{0.0057}\\
Plant 3 & 0.1820 & - & - & 0.0062 & \textbf{0.0053} & 0.0169 & 0.0057\\
Plant 4 & 0.1608 & - & - & 0.0075 & \textbf{0.0050} & 0.0169 & 0.0056\\
Plant 5 & 0.1892 & - & 0.3111 & 0.0079 & \textbf{0.0052} & 0.0359 & 0.0056\\
Plant 6 & 0.1347 & - & - & 0.0087 & 0.0069 & 0.0152 & \textbf{0.0062}\\
Plant 7 & 0.1032 & - & - & 0.0114 & \textbf{0.0056} & 0.0279 & 0.0084\\
Plant 8 & 0.2052 & - & 0.3214 & 0.0075 & - & 0.0261 & \textbf{0.0058}\\
\midrule
Mean    & 0.1674 & - & 0.3170 & 0.0083 & \textbf{0.0056} & 0.0216 & 0.0065\\
Std.\,Dev. & 0.0337 & - & 0.0043 & 0.0015 & \textbf{0.0007} & 0.0072 & 0.0013\\
\bottomrule
\end{tabular}
\end{table}

\begin{table}[t!]
\centering
\caption{Comparison of two-sided HD (reconstruction and ground truth point cloud) for eight different plants.}
\label{tab:plants_two_sided_hd_pred_gt}
\scriptsize
\setlength{\tabcolsep}{3pt}
\begin{tabular}{@{}l@{\hspace{4pt}}cccccccc@{}} 
\toprule
Model & PSR & NP & IGR & DiGS & OG-INR & NSH & STITCH (Ours) \\
\midrule
Plant 1 & 0.7838 & - & 0.9290 & \textbf{0.0380} & - & 0.0855 & 0.1005\\
Plant 2 & 0.6917 & 0.1838 & - & 0.0426 & - & 0.1581 & \textbf{0.0257}\\
Plant 3 & 0.7258 & - & - & 0.0351 & 0.0225 & 0.1218 & \textbf{0.0208}\\
Plant 4 & 0.8691 & - & - & 0.0457 & 0.0221 & 0.1078 & \textbf{0.0216}\\
Plant 5 & 0.6862 & - & 0.5423 & 0.0403 & 0.0276 & 0.1788 & \textbf{0.0178}\\
Plant 6 & 0.6918 & - & - & 0.0457 & 0.0464 & 0.1047 & \textbf{0.0264}\\
Plant 7 & 0.7878 & - & - & 0.0721 & \textbf{0.0314} & 0.1941 & 0.0680\\
Plant 8 & 0.8263 & - & 0.8302 & 0.0374 & - & 0.1580 & \textbf{0.0264}\\
\midrule
Mean    & 0.7578 & - & 0.7672 & 0.0446 & \textbf{0.0300} & 0.1386 & 0.0384\\
Std.\,Dev. & 0.0647 & - & 0.1640 & 0.0110 & \textbf{0.0089} & 0.0365 & 0.0278\\
\bottomrule
\end{tabular}
\end{table}

\end{document}